\documentclass[fullpaper,final]{nldl}

\paperID{45}

\vol{352}

\usepackage{mathtools}

\usepackage{graphicx}
\usepackage{array}

\usepackage{booktabs}

\usepackage{enumitem}

\usepackage{algorithm}
\usepackage{algorithmic}

\usepackage{listings}
\usepackage{hyperref}
\usepackage{url}
\hypersetup{
  pdfusetitle,
  colorlinks,
  linkcolor = BrickRed,
  citecolor = NavyBlue,
  urlcolor  = Magenta!80!black,
}

\newcommand{\tablestyle}{\footnotesize\setlength{\tabcolsep}{4pt}\renewcommand{\arraystretch}{1.1}}
\newcommand{\qlabel}[1]{\rotatebox{90}{\parbox[c]{\qimg}{\centering\scriptsize\hspace{0pt}#1}}}
\newlength{\qimg}

\title{CrevasseSeg: A Label-Efficient UAV Crevasse Segmentation Framework}
\author[1,2]{Steven Wallace\thanks{Corresponding Author.}}
\author[2,3]{William D. Harcourt}
\author[4]{Richard Hann}
\author[5]{Aiden Durrant}
\author[1]{Somayajulu Sripada}
\author[6]{Georgios Leontidis}

\affil[1]{School of Natural and Computing Sciences, University of Aberdeen, UK}
\affil[2]{Interdisciplinary Institute, University of Aberdeen, UK}
\affil[3]{School of Geosciences, University of Aberdeen, UK}
\affil[4]{Department of Engineering Cybernetics, Norwegian University of Science and Technology, Norway}
\affil[5]{School of Computing Sciences, University of East Anglia, UK}
\affil[6]{Department of Physics and Technology, UiT The Arctic University of Norway}

\begin{document}
\maketitle

\begin{abstract}

Crevasse mapping from uncrewed aerial vehicle (UAV) imagery supports glaciological research and field safety, but pixel-level annotation requires costly expert effort. We introduce \textbf{CrevasseSeg}, a binary segmentation evaluation framework from the terminus of Borebreen, Svalbard, comprising 2{,}244 unlabelled UAV orthomosaic tiles for self-supervised/unsupervised fine-tuning and 200 labelled tiles (24 validation, 176 test). We benchmark five self-supervised objectives (BYOL, Jensen-Shannon Divergence (JSD), Barlow Twins, VICReg and a combined BYOL–JSD) across O-Net, O-Net++ and a DINOv3-initialised O-Net. Frozen features are read out with a linear probe and an XGBoost classifier, both fit only on the 24 validation tiles. At the final decoder block, the two readouts invert. DINOv3 models are mostly the weakest under linear probing, often collapsing to a constant prediction, yet they are the strongest under XGBoost. This pattern does not hold at earlier decoder blocks. UMAP suggests why: DINOv3 fragments pixels into many small, locally interleaved clusters, whereas the convolutional models embed them on a single class-sorted manifold. Satellite-pretrained DINOv3 beats natural-image initialisation for every objective under XGBoost (+9.7 mDSC on average), but not consistently under linear probing. Our pipeline, DINOv3-ViT-L-Sat-O-Net-BYOL-JSD, reaches 74.75 mDSC / 61.16 mIoU (77.04 / 64.28 with a decoder-block ensemble), on par with BYOL alone, and outperforms RGB-pixel ML baselines. However, a frozen DINOv3-Sat encoder with an XGBoost readout alone reaches 77.18 / 64.17. It matches or exceeds every SSL variant and comes within 0.71 mDSC of a fully supervised DINOv3 U-Net, highlighting satellite foundation models as a strong baseline. We release CrevasseSeg as a benchmark for label-efficient segmentation in remote sensing. Our code and dataset are available at: \href{https://github.com/Stevieee83/CrevasseSeg-a-label-efficient-uav-crevasse-segmentation-framework}{CrevasseSeg}

\end{abstract}

\section{Introduction}

Self-supervised learning (SSL) has become a central paradigm in Computer Vision, enabling deep learning networks to learn transferable representations from unlabelled data \cite{chen2020simple, he2020momentum, grill2020bootstrap, zbontar2021barlow, bardes2021vicreg, assran2023self}. Self-supervised learning is most appealing when supervised learning is restricted by the quantity of annotated labels, which is labour-intensive, demands expert judgement and does not scale.  Semantic segmentation is the extreme case, because every pixel must be labelled and, in scientific applications, verified by a domain expert. In remote sensing applications, producing thousands of densely annotated images to fully supervise pipelines is infeasible or prohibitively expensive.

Crevasses on glaciers, which are indicators of ice dynamics and represent a significant danger during polar fieldwork, are a domain in which automated mapping has only recently been developed \cite{foroutan2019automatic, surawy2023mapping}. Whilst they have traditionally been mapped in satellite imagery by hand \cite{sevestre2018tidewater}, the proliferation of UAVs and the ability to capture centimetre-scale orthomosaics of glacier surfaces using structure-from-motion surveys \cite{ryan2015uav, chudley2019high} makes automated segmentation attractive. However, the labelled data required to train segmentation models in this setting barely exists.

In this work, we contribute a benchmark, not a single model. We assemble CrevasseSeg from UAV orthomosaic imagery of a highly crevassed tidewater glacier terminus in Svalbard (named Borebreen), and use it to ask a focused question: \emph{how far can self-supervised representations, combined with a small labelled dataset, go on crevasse segmentation?} We evaluate a matrix of SSL objectives and segmentation architectures, including the DINOv3 \cite{simeoni2025dinov3} foundation model initialisation under two feature readouts that differ only in whether the decision boundary is linear or non-linear.

Our contributions are:

\begin{enumerate}[itemsep=2pt, topsep=2pt]
    \item \textbf{CrevasseSeg}, a new public benchmark dataset for binary crevasse segmentation from UAV orthomosaic imagery, with a fixed 2{,}244 / 24 / 176 train–validation–test split in which only 24 labelled images are available for fitting any classifier, and in which splits are separated by a buffer set from the measured spatial autocorrelation range of the imagery.
    \item A systematic \textbf{evaluation framework} of five SSL objectives (BYOL, JSD, Barlow Twins, VICReg, BYOL–JSD) across three architectures (O-Net, O-Net++, DINOv3-O-Net), with both a linear probe and an XGBoost readout.
    \item An \textbf{analysis of how the feature readout interacts with the representation}: we probe each frozen SSL backbone with a linear and a non-linear classifier, layer by layer, and use UMAP visualisations of the learned feature space to characterise when foundation-model-based features are linearly separable and when they are not.
    \item A \textbf{label-efficient pipeline} that fine-tunes a DINOv3-initialised O-Net with self-supervision and reads out crevasse masks from only 24 labelled images using an XGBoost classifier and a decoder-block soft-voting ensemble, benchmarked against standard machine-learning classifiers, recent unsupervised-segmentation methods and a frozen-encoder baseline.
\end{enumerate}

\section{Related Work}

\subsection{Self-supervised and unsupervised Segmentation}

A large family of SSL methods learn representations by enforcing consistency between augmented views without labels, including contrastive \cite{chen2020simple, he2020momentum}, distillation-based \cite{grill2020bootstrap, chen2021exploring}, and redundancy-reduction \cite{zbontar2021barlow, bardes2021vicreg} objectives. However, most of these methods were originally developed for image classification. Fully supervised segmentation by contrast relies on dense annotation \cite{long2015fully, ronneberger2015u, zhou2018unet++, badrinarayanan2015segnet, chen2017deeplab, chen2018encoder, zhao2017pyramid, xie2021segformer}, which is precisely the issue that this work seeks to avoid.

Unsupervised semantic segmentation targets this gap. Classical pipelines combined clustering with superpixel over-segmentation such as SLIC \cite{achanta2012slic}. Deep learning approaches advanced the field; PiCIE \cite{cho2021picie} clusters pixel features under photometric-invariance and geometric-equivalence biases, while STEGO \cite{hamilton2022unsupervised} distils the semantically consistent correspondences already present in self-supervised ViT features \cite{caron2021emerging} into discrete labels. Building on STEGO’s correspondence-distillation framework, several methods refine it along different axes: HP \cite{seong2023leveraging} mines global and local hidden positives to strengthen semantic and spatial consistency, SmooSeg \cite{lan2023smooseg} imposes a smoothness prior that promotes piecewise-coherent segments while preserving boundaries between them, EAGLE \cite{kim2024eagle} introduces object-centric spectral cues (via graph Laplacian eigenvectors) to better bind together the disparate parts of a single object and PriMaPs-EM \cite{Hahn:2024:BUS} decomposes images into principal mask proposals and fits class prototypes to them with a stochastic Expectation-Maximization (EM) algorithm.

In remote sensing, where labels are scarce and domain shift is present, unsupervised single-scene segmentation \cite{saha2022unsupervised} and unsupervised domain adaptation across sensors \cite{zhu2023unsupervised} have been proposed. These directions are especially relevant to UAV imagery where rapid deployment over previously unmapped terrain means labelled data are rarely available in advance.

\subsection{Foundation models, BYOL and JSD}

BYOL \cite{grill2020bootstrap} trains an online network to predict the projection of a second, augmented view generated by a momentum-updated target network, learning representations without labels or negative pairs. The O-Net architecture \cite{zhou2025onet} instead enforces label-free consistency between the predictive distributions of a Siamese U-Net's two branches via a Jensen-Shannon Divergence (JSD) loss. As BYOL enforces consistency across augmented views while JSD enforces consistency across predicted distributions, combining the two offers a potentially complementary self-supervisory signal, motivating the proposed method evaluated in this work. We additionally benchmark DINOv3 \cite{simeoni2025dinov3}, extending the DINO self-supervised line \cite{caron2021emerging, oquab2023dinov2}, as a frozen encoder within a convolutional decoder (DINOv3-O-Net).

\subsection{Crevasse mapping and UAV surveys}

The spatial distribution and orientation of glacier crevasses reflect ice deformation as a glacier flows \cite{colgan2016glacier}; hence, mapping crevasses can improve our understanding of glacier stress-strain relationships. Furthermore, icebergs detach from the end of tidewater glaciers where crevasses extend the full ice thickness at any particular locality \cite{benn2026calving}, which is a critical component of mass loss and sea level changes. Operationally, crevasse maps underpin safe route planning for polar fieldwork and glacier travel \cite{lever2013autonomous}. Synthetic Aperture Radar (SAR) data has improved detection of surface and buried crevasses \cite{vaughan1993relating, marsh2021crevasse}, but glaciers are continuously evolving, and hence the need for up-to-date inventories grows. UAV photogrammetry can deliver centimetre-scale Digital Elevation Models (DEMs) and orthomosaics of glacier surfaces \cite{ryan2015uav, chudley2019high, fugazza2015high}; DEMs support morphometric analysis while orthomosaics enable image-based fracture detection that bypasses terrain interpolation artefacts in densely fractured areas. UAV surveys may also be repeatable at sub-weekly intervals, capturing rapid crevasse-field evolution \cite{baurley2022assessing}. Svalbard in particular has seen rapid growth in UAV-based environmental and glaciological research in recent years \cite{hann2026sess}. This paper targets crevasse detection directly from UAV orthomosaic imagery.

\section{The CrevasseSeg benchmark}

\subsection{Region of interest and acquisition}

CrevasseSeg covers part of the highly crevassed terminus of \textbf{Borebreen}, a tidewater glacier in Svalbard. The regions of interest taken from Borebreen are from two years 8th August 2023 \cite{B553MB_2024} and 8th September 2025 \cite{ZAJLVT_2025}, spanning 0.412 km$^2$ and 1.182 km$^2$ of the terminus, respectively (\autoref{fig:aoi-both}). The terminus was prioritised because it is the region most susceptible to iceberg calving. Imagery was acquired by a DJI Mavic 2 Pro Enterprise UAV in 2023 and a Mavic 3 Pro Enterprise UAV in 2025, and extracted from orthomosaics that were tiled into non-overlapping $512\times512$ resolution patches.

\subsection{Annotation and splits}

Test and validation tiles were annotated at pixel level as a binary mask \emph{crevasse} (foreground) versus \emph{ice} (background) by a single annotator and another expert approver using the Computer Vision Annotation Tool (CVAT) software. The benchmark uses a single fixed split (\autoref{tab:split}). Critically, \textbf{only the 24 validation images carry labels available for fitting any classifier}: The 2{,}244 training tiles (1180 from 2023 and 1064 from 2025) are unlabelled and used solely for self-supervised/unsupervised pre-training or fine-tuning, and the 176 labelled test tiles are held out for model evaluation. This makes CrevasseSeg an explicit low-label benchmark rather than a conventional supervised split.

\begin{table}[tb]
  \centering
  \caption{CrevasseSeg split. Labels are available only for the validation and test sets.}
  \label{tab:split}
  \begin{tabular}{@{}lccc@{}}
    \toprule
    Split & \#Tiles & Labelled & Role \\
    \midrule
    Train & 2{,}244 & No  & Unlabelled fine-tune \\
    Val   & 24      & Yes & Fit readout / tune \\
    Test  & 176     & Yes & Evaluation \\
    \bottomrule
  \end{tabular}
\end{table}

\subsection{Spatially buffered splits}
\label{sec:buffer}

Neighbouring tiles cut from the same orthomosaic share texture, illumination and crevasse patterns, so placing adjacent tiles in different splits leaks information across them and inflates test scores \cite{roberts2017cross, ploton2020spatial}. We therefore assign whole, spatially contiguous ROIs to each split and enforce a minimum separation between ROIs of different splits, set from the measured spatial autocorrelation range of the imagery rather than chosen by hand.

The orthomosaic was projected to UTM zone 33N (EPSG:32633; 3.54\,cm pixels), and eight $20{,}000\times20{,}000$ pixel windows ($\approx$708\,m $\times$ 708\,m) were sampled across it with a fixed random seed. In each window $w$ we computed the normalised 2-D autocorrelation $\rho_w$ of the image via the FFT (Wiener–Khinchin theorem) and extracted directional profiles $\rho_w(h,\theta)$ up to a lag of 8{,}000 pixels ($\approx$283\,m). The range of a profile is the smallest lag at which it falls below $e^{-1}$; each window's range $r_w$ is taken along its most persistent orientation (the slowest-decaying direction), and profiles were also checked for periodic side-lobes (minimum prominence 0.1) that would indicate regular crevasse spacing; none were detected. The buffer is then
\begin{equation}
  d_{\text{buf}} = k \cdot \max_{w} r_w + \sigma_{\text{pos}},
\end{equation}
with safety multiplier $k=1.5$ and positional uncertainty $\sigma_{\text{pos}}=0$, since all tiles of a survey are cut from a single co-registered orthomosaic.

\begin{figure}[tb]
  \centering
  \includegraphics[width=\columnwidth]{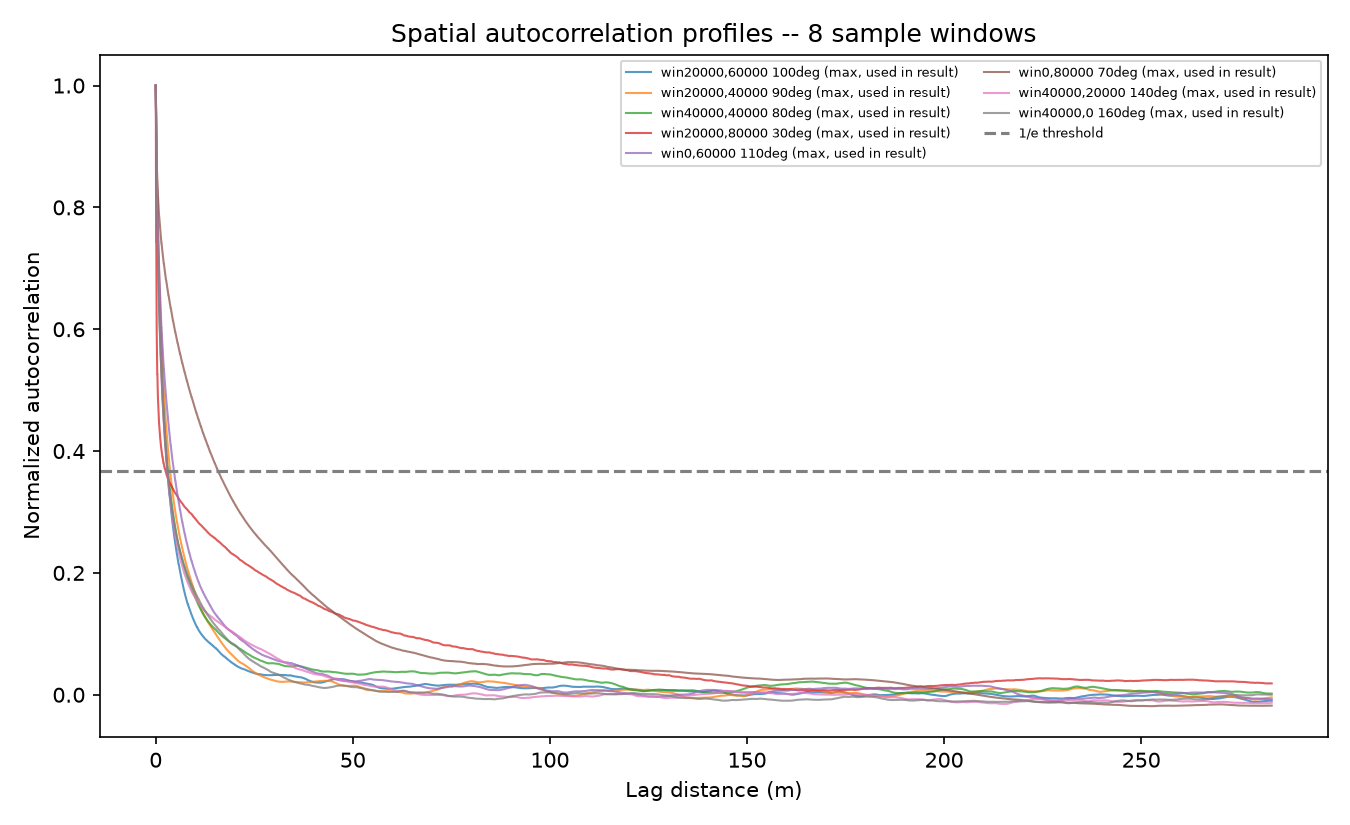}
  \caption{Directional spatial autocorrelation profiles (most persistent orientation) for the eight sample windows of the 2023 orthomosaic. The dashed line marks the $1/e$ threshold defining the range. The maximum range (15.79\,m) $\times\,1.5$ gives the 23.68\,m buffer enforced between splits.}
  \label{fig:autocorr}
\end{figure}

\autoref{fig:autocorr} shows the resulting profiles. Seven windows decorrelate within 2.5–4.6\,m, while one (70$^\circ$ orientation) decays far more slowly, with a range of 15.79\,m. Taking the maximum rather than the median window range (3.04\,m) is deliberately conservative and yields $d_{\text{buf}}=23.68$\,m, i.e.\ $\approx$669 pixels or $\approx$1.3 tile widths (one $512$-pixel tile spans 18.1\,m). At this distance every profile has decayed below $\approx$0.3. This buffer is enforced between the training, validation and test ROIs (\autoref{fig:aoi-both}), and all results in this paper are obtained on the buffered splits.

\subsection{Evaluation framework}

We report foreground mean Dice Score Similarity (mDSC) and mean Intersection over Union (mIoU) on the 176 test tiles. All segmentation models are first pre-trained/fine-tuned with SSL or unsupervised learning on the unlabelled train split. All models that use a DINO/DINOv3 image encoder for feature extraction are frozen, where only the convolutional decoder is trained. The frozen per-pixel features taken from the output of the final (third) decoder block unless stated otherwise, are then read out by a classifier trained on 24 labelled validation tiles. We evaluate two readouts that differ only in the form of the decision boundary:

\begin{enumerate}[label=(\roman*), itemsep=1pt, topsep=2pt]
  \item a \textbf{linear} probe, and
  \item a non-linear \textbf{XGBoost} gradient-boosted-tree classifier.
\end{enumerate}

\noindent Because most rankings among the strongest methods are separated by small margins, we recommend reporting multi-seed variance; we return to this in \autoref{sec:limitations}.

\begin{figure*}[tb]
  \centering
  \includegraphics[width=0.95\textwidth]{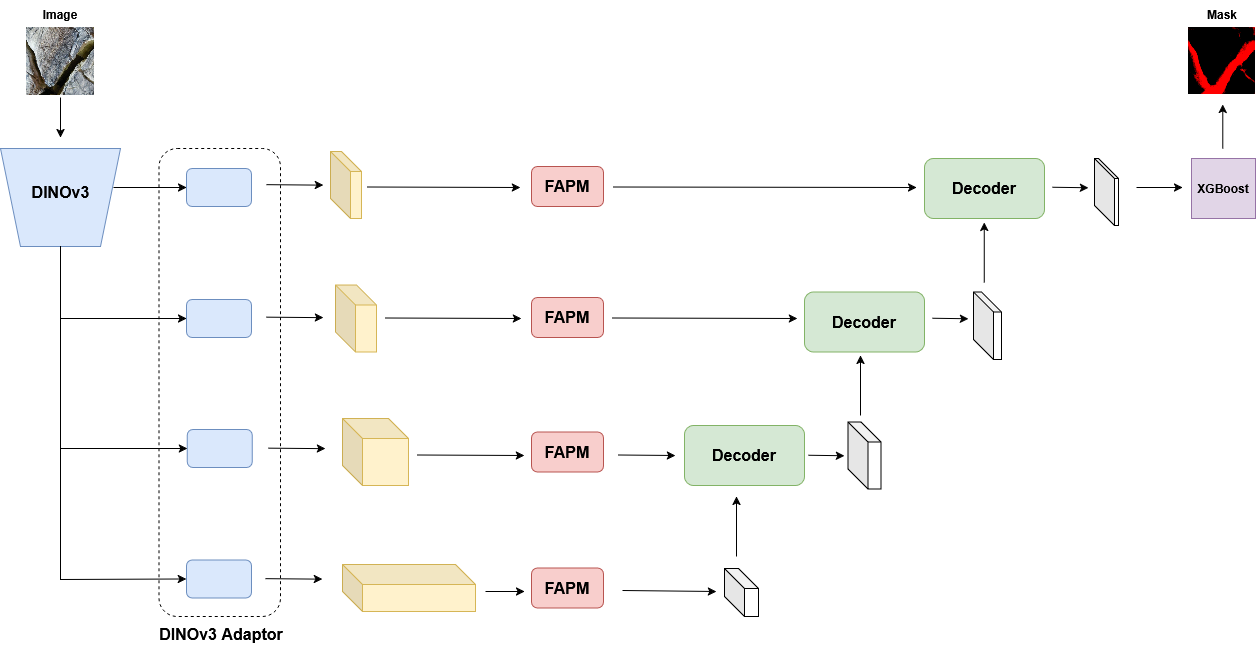}
  \caption{Overview of the DINOv3-O-Net architecture. The frozen DINOv3 backbone extracts multi-scale features, which are passed through per-block adaptor layers (DINOv3 Adaptor) and a Feature Alignment/Pyramid Module (FAPM) before being progressively fused by a stack of decoders to produce the final binary crevasse segmentation mask output from XGBoost.}
  \label{fig:architecture}
\end{figure*}

\section{Framework and benchmark methods}

\subsection{Architectures}

We compare three segmentation architectures under an identical SSL and readout protocol. \textbf{O-Net} is a Siamese U-Net \cite{zhou2025onet}; \textbf{O-Net++} replaces its encoder–decoder with a nested U-Net++ architecture \cite{zhou2018unet++}; and \textbf{DINOv3-O-Net} initialises the encoder with a DINOv3 ViT-L backbone \cite{simeoni2025dinov3, xue2025dinov3}, using either the natural-image checkpoint (\emph{DINOv3-VIT-L}) or the satellite-pre-trained checkpoint (\emph{DINOv3-ViT-L-Sat}).

\autoref{fig:architecture} illustrates the DINOv3-O-Net design. The frozen DINOv3 ViT-L backbone produces multi-scale token features that are passed through lightweight per-scale adaptor layers (the \emph{DINOv3} Adaptor) and the Feature Alignment and Pyramid Module (FAPM), after which a cascade of decoders progressively fuses them from coarse to fine, producing the features used to generate the binary crevasse mask output from a further linear probe or XGBoost classifier. Only the adaptor, FAPM and decoder parameters are adapted during SSL fine-tuning and readout; the DINOv3 backbone remains frozen throughout, consistent with the frozen-encoder protocol used for both DINOv3-O-Net checkpoint variants.

\subsection{Self-supervised objectives}

Each model is fine-tuned with each of the five objectives. \textbf{BYOL} \cite{grill2020bootstrap} minimises the normalised prediction error between an online prediction $q(z_\theta)$ of one view and the target projection $z’_\xi$ of a second view produced by a momentum encoder.

\begin{equation}
  \mathcal{L}_{\text{BYOL}} = 2 - 2\cdot\frac{\langle q(z_\theta),\, z'_\xi\rangle}{\lVert q(z_\theta)\rVert\, \lVert z'_\xi\rVert}.
\end{equation}
The \textbf{JSD} objective \cite{zhou2025onet} enforces consistency between the two Siamese branches' segmentation distributions $P$ and $Q$:
\begin{equation}
  \mathcal{L}_{\text{JSD}} = \tfrac{1}{2}\mathrm{KL}(P\Vert M) + \tfrac{1}{2}\mathrm{KL}(Q\Vert M),\quad M=\tfrac{1}{2}(P+Q).
\end{equation}
\textbf{Barlow Twins} \cite{zbontar2021barlow} and \textbf{VICReg} \cite{bardes2021vicreg} provide redundancy-reduction and variance--invariance--covariance baselines respectively. Our combined \textbf{BYOL--JSD} objective sums the two:
\begin{equation}
  \mathcal{L} = \mathcal{L}_{\text{BYOL}} + \lambda\,\mathcal{L}_{\text{JSD}},
\end{equation}

with $\lambda=0.5$ balancing the terms. To fine-tune each architecture with each SSL method (decoder only for DINOv3-O-Net; end-to-end for O-Net and O-Net++), we used the AdamW optimiser and a Reduce-on-Plateau learning rate scheduler, with a per-GPU batch size of 256 under PyTorch's Distributed Data Parallel (DDP), giving a total batch size of 768, training for 500 epochs on 3 NVIDIA A100 GPUs and retaining the checkpoint with the lowest SSL loss. Since this is a segmentation task, no augmentations that alter the spatial orientation of the input image were used; we applied only photometric augmentations — random colour jittering, random Gaussian blurring, and random greyscaling.

\subsection{Label-efficient readout}

After SSL fine-tuning, per-pixel features are extracted, and a classifier is fit on the 24 labelled validation tiles to predict the two classes. We tune both the linear probe and the XGBoost hyperparameters on validation. Our final system, \textbf{DINOv3-ViT-L-Sat-O-Net-BYOL-JSD}, uses the satellite-pretrained DINOv3-O-Net features with an XGBoost readout to classify each pixel as crevasse (foreground) or ice (background).

\section{Results}

\subsection{Linear vs.\ non-linear readout}

\autoref{tab:ssl} reports the five SSL objectives for the four backbones under both readouts, and the two readouts give contrasting pictures. Under \emph{linear} evaluation, the convolutional backbones give the best result for four of the five objectives, with O-Net++-BYOL reaching the highest linear score (70.21 mDSC / 55.26 mIoU). Most DINOv3 rows sit at 48.27 mDSC / 33.87 mIoU, the score of a constant single-class prediction, i.e.\ the linear probe fails to separate crevasse from ice. Only the satellite checkpoint rises above this level, and only modestly: DINOv3-Sat-Barlow-Twins (54.30 mDSC / 39.02 mIoU, the best linear result for that objective) and DINOv3-Sat-JSD (50.70 mDSC / 35.87 mIoU).

Under the \emph{non-linear} XGBoost readout, the ranking inverts. The satellite-pretrained DINOv3-O-Net becomes the strongest backbone for all five objectives (65.59–74.78 mDSC), the natural-image DINOv3-O-Net is second for BYOL-JSD and BYOL, and no DINOv3 row remains at the collapse level. In summary, \textbf{the final-decoder-block features of the DINOv3 models encode crevasse structure that a linear boundary largely cannot recover, but a non-linear readout can}. Practically, foundation-model-based features on this task should not be judged by linear probing alone.

\begin{table*}[tb]
\centering
\tablestyle
\caption{Five SSL objectives $\times$ four backbones on CrevasseSeg test (foreground mDSC / mIoU), under the linear probe and the XGBoost readout. Best per objective and readout in \textbf{bold}.}
\label{tab:ssl}
\begin{tabular}{@{}l rr @{\hspace{18pt}} rr@{}}
\toprule
 & \multicolumn{2}{c}{Linear} & \multicolumn{2}{c}{XGBoost} \\
\cmidrule(lr){2-3}\cmidrule(l){4-5}
Methods  & mDSC  & mIoU & mDSC & mIoU \\
\midrule
DINOv3-ViT-L-Sat-O-Net-BYOL-JSD     & 48.27 & 33.87                   & \textbf{74.75} & \textbf{61.16} \\
DINOv3-ViT-L-O-Net-BYOL-JSD         & 48.82 & 34.33                   & 73.84 & 59.71 \\
O-Net-BYOL-JSD                      & 48.67 & 33.64                   & 38.79 & 25.52 \\
O-Net-Plus-Plus-BYOL-JSD            & \textbf{51.95} & \textbf{36.62} & 39.22 & 25.89 \\
\midrule
DINOv3-ViT-L-Sat-O-Net-BYOL         & 48.27 & 33.87                   & \textbf{74.78}  & \textbf{61.28}\\
DINOv3-ViT-L-O-Net-BYOL             & 48.48 & 33.97                   & 72.91 & 58.49                   \\
O-Net-BYOL                          & 68.59 & 53.30                   & 54.29 & 38.57                   \\
O-Net-Plus-Plus-BYOL                & \textbf{70.21} & \textbf{55.26} & 60.90 & 44.98 \\
\midrule
DINOv3-ViT-L-Sat-O-Net-JSD          & 50.70 & 35.87                   & \textbf{65.59}  & \textbf{50.57} \\
DINOv3-ViT-L-O-Net-JSD              & 48.28 & 33.87                   & 52.71 & 37.22 \\
O-Net-JSD                           & \textbf{66.98} & \textbf{51.26} & 61.28 & 45.12 \\
O-Net-Plus-Plus-JSD                 & 65.96 & 50.39                   & 59.88 & 43.84 \\
\midrule
DINOv3-ViT-L-Sat-O-Net-Barlow-Twins & \textbf{54.30} & \textbf{39.02} & \textbf{72.77} & \textbf{58.57} \\
DINOv3-ViT-L-O-Net-Barlow-Twins     & 48.27 & 33.87                   & 56.54 & 40.50 \\
O-Net-Barlow-Twins                  & 48.21 & 33.73                   & 57.88 & 41.69 \\
O-Net-Plus-Plus-Barlow-Twins        & 52.86 & 36.99                   & 59.51 & 43.13 \\
\midrule
DINOv3-ViT-L-Sat-O-Net-VICReg       & 48.27 & 33.87                   & \textbf{73.50} & \textbf{59.60} \\
DINOv3-ViT-L-O-Net-VICReg           & 48.27 & 33.86                   & 57.04 & 41.03 \\
O-Net-VICReg                        & \textbf{54.03} & \textbf{38.41} & 67.54 & 52.20 \\
O-Net-Plus-Plus-VICReg              & 49.94 & 35.33                   & 56.70 & 41.01 \\
\bottomrule
\end{tabular}
\end{table*}

\subsection{Effect of satellite pre-training}

Under the XGBoost readout (\autoref{tab:ssl}), the satellite-pretrained DINOv3 variant improves over its natural-image counterpart for every objective, by an average of +9.67 mDSC / +10.85 mIoU. Gains are small for the BYOL-based objectives (+0.91 mDSC for BYOL-JSD, +1.87 mDSC for BYOL) and large for JSD (+12.88 mDSC), Barlow Twins (+16.23 mDSC) and VICReg (+16.46 mDSC). This suggests domain-matched pre-training transfers meaningfully to UAV imagery of glaciers, consistent with the benefits reported for foundation models elsewhere in remote sensing. Under linear probing the picture is mixed: the satellite checkpoint helps markedly for Barlow Twins (+6.03 mDSC) and slightly for JSD (+2.42 mDSC), but both checkpoints collapse for BYOL-JSD, BYOL and VICReg.

\subsection{Effect of SSL objective}

On the satellite backbone under XGBoost, BYOL (74.78 mDSC / 61.28 mIoU) and BYOL-JSD (74.75 mDSC / 61.16 mIoU) are effectively tied, followed by VICReg (73.50 mDSC / 59.60 mIoU) and Barlow Twins (72.77 mDSC / 58.57 mIoU), with JSD alone clearly weaker (65.59 mDSC / 50.57 mIoU). On the natural-image checkpoint, BYOL-JSD leads BYOL (73.84 vs.\ 72.91 mDSC). These margins are about one point or less, which a single-seed evaluation cannot resolve, so we do not claim that the JSD term contributes to performance; the BYOL term appears to be the main driver. Moreover, any benefit is confined to the frozen-encoder setting: when the encoder is trained end-to-end, adding JSD to BYOL degrades O-Net and O-Net++ sharply under both readouts (XGBoost: 54.29$\rightarrow$38.79 and 60.90$\rightarrow$39.22 mDSC; linear: 68.59$\rightarrow$48.67 and 70.21$\rightarrow$51.95 mDSC). A plausible explanation is that with a trainable encoder, the cross-branch consistency term can be satisfied by less informative representations, whereas a frozen pre-trained encoder prevents this. We therefore frame BYOL-JSD as a competitive objective for frozen-encoder fine-tuning, on par with BYOL, not as a generally superior one.

\subsection{Comparison with shared baselines}

\autoref{tab:ml-baselines} compares our pipeline against standard machine-learning classifiers using raw RGB pixel values, fit on the same 24 labelled images and evaluated on the same test set. Our single-model DINOv3-ViT-L-Sat-O-Net-BYOL-JSD with an XGBoost readout (74.75 mDSC / 61.16 mIoU) outperforms all of them. The supervised raw-pixel classifiers (Random Forest, XGBoost, KNN) score 47.39–48.22 mDSC, close to the constant-prediction level, and the FNN and LSTM score lower still; the strongest baseline is the unsupervised K-Means (70.07 mDSC / 55.02 mIoU), followed by GMM (57.01 mDSC / 41.37 mIoU). Because the XGBoost classifier and the 24 training labels are identical in our pipeline and in the raw-pixel XGBoost baseline (47.39 mDSC / 32.67 mIoU), the +27.36 mDSC / +28.49 mIoU margin between them is attributable to the features rather than to the classifier.

\begin{table}[tb]
  \caption{CrevasseSeg test: our label-efficient pipeline vs.\ standard baselines fit on the same 24 labels. Best in \textbf{bold}.}
  \label{tab:ml-baselines}
  \centering
  \tablestyle\setlength{\tabcolsep}{1.5pt}
  \begin{tabular*}{\columnwidth}{@{}l@{\extracolsep{\fill}}rr@{}}
    \toprule
    Method & mDSC & mIoU\\
    \midrule
    DINOv3-ViT-L-Sat-O-Net-BYOL-JSD (ours)     & \textbf{74.75} & \textbf{61.16} \\
    Random Forest                              & 47.50 & 32.74   \\
    XGBoost                                    & 47.39 & 32.67   \\
    FNN                                        & 34.09 & 21.93  \\
    LSTM                                       & 37.23 & 24.27   \\
    KNN                                        & 48.22 & 33.34  \\
    K-Means                                    & 70.07 & 55.02   \\
    GMM                                        & 57.01 & 41.37   \\
  \bottomrule
  \end{tabular*}
\end{table}

\begin{table*}[tb]
  \caption{CrevasseSeg test: our label-efficient pipeline vs.\ unsupervised segmentation baselines under linear and XGBoost readouts. Best per readout in \textbf{bold}.}
  \label{tab:dl}
  \centering
  \tablestyle
  \begin{tabular}{@{}l rr @{\hspace{18pt}} rr@{}}
    \toprule
     & \multicolumn{2}{c}{Linear} & \multicolumn{2}{c}{XGBoost} \\
    \cmidrule(lr){2-3}\cmidrule(l){4-5}
    Method & mDSC & mIoU & mDSC & mIoU \\
    \midrule
    DINOv3-ViT-L-Sat-O-Net-BYOL-JSD-Ensemble (ours) & 74.09 & 60.34                    & 77.04          & 64.28 \\
    DINOv3-ViT-L-Sat-O-Net-BYOL-JSD (ours)          & 48.27 & 33.87                    & 74.75          & 61.16 \\
    PriMaPs-EM-DINO-ViT-B-16                        & 73.22 & 59.19                    & 73.70          & 59.78 \\
    EAGLE-DINO-ViT-B-8                              & \textbf{76.32} & \textbf{63.08}  & 77.41          & 64.49 \\
    HP-DINO-ViT-S-8                                 & 75.79 & 62.31                    & \textbf{78.54} & \textbf{66.10} \\
    STEGO-DINO-ViT-B-8                              & 75.24 & 61.90                    & 76.37          & 63.31 \\
    SmooSeg-DINO-ViT-S-8                            & 73.92 & 60.40                    & 74.61          & 61.13 \\
  \bottomrule
  \end{tabular}
\end{table*}

We further compare against recent unsupervised semantic segmentation methods — PriMaPs-EM \cite{Hahn:2024:BUS}, EAGLE \cite{kim2024eagle}, HP \cite{seong2023leveraging}, STEGO \cite{hamilton2022unsupervised} and SmooSeg \cite{lan2023smooseg} — under both readouts (\autoref{tab:dl}). Alongside the single model, we report a soft-voting ensemble that averages the readout class probabilities of the three decoder-block outputs of the same backbone (upsampled to label resolution). Under the non-linear readout, the single model exceeds PriMaPs-EM and SmooSeg but trails STEGO, EAGLE and HP. The ensemble closes most of this gap, reaching 77.04 mDSC / 64.28 mIoU: essentially level with EAGLE and 1.50 mDSC behind the best method, HP. Under linear probing, the single model collapses, whereas the ensemble recovers to 74.09 mDSC / 60.34 mIoU, marginally above PriMaPs-EM and SmooSeg, below STEGO and HP, and 2.23 mDSC behind the best method, EAGLE. Several of these margins are around one point and cannot be resolved without multi-seed evaluation (\autoref{sec:limitations}). We therefore frame the SSL-plus-ensemble pipeline as competitive with dedicated unsupervised segmentation models under a non-linear readout while remaining highly label-efficient, and flag the remaining linear-probe gap as a target for future work.

\begin{figure*}[tb]
  \centering
  \begin{subfigure}[b]{0.35\textwidth}
    \includegraphics[width=\linewidth]{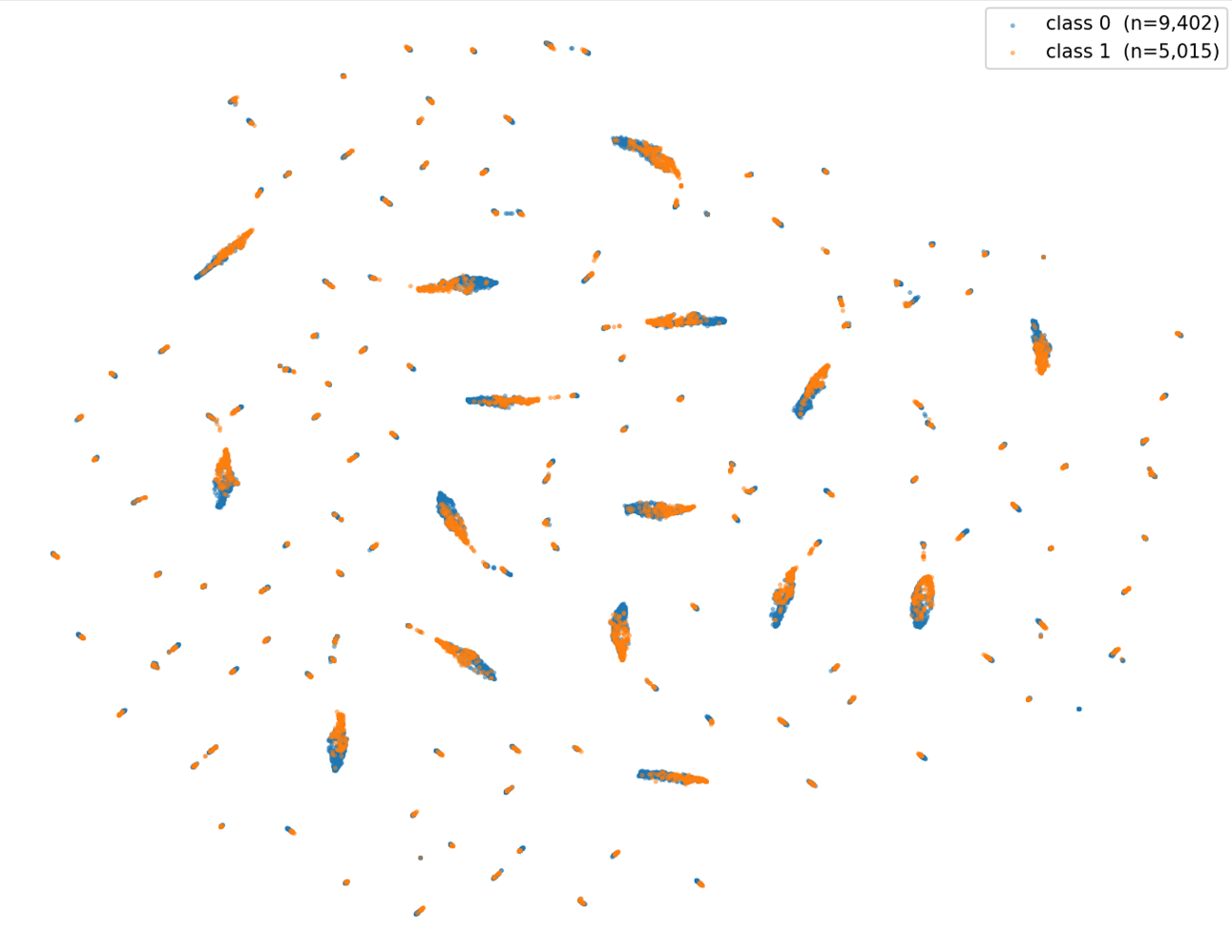}
    \caption{DINOv3-ViT-L-Sat-O-Net}
    \label{fig:umap-sat}
  \end{subfigure}
  \hspace{0.05\textwidth}
  \begin{subfigure}[b]{0.35\textwidth}
    \includegraphics[width=\linewidth]{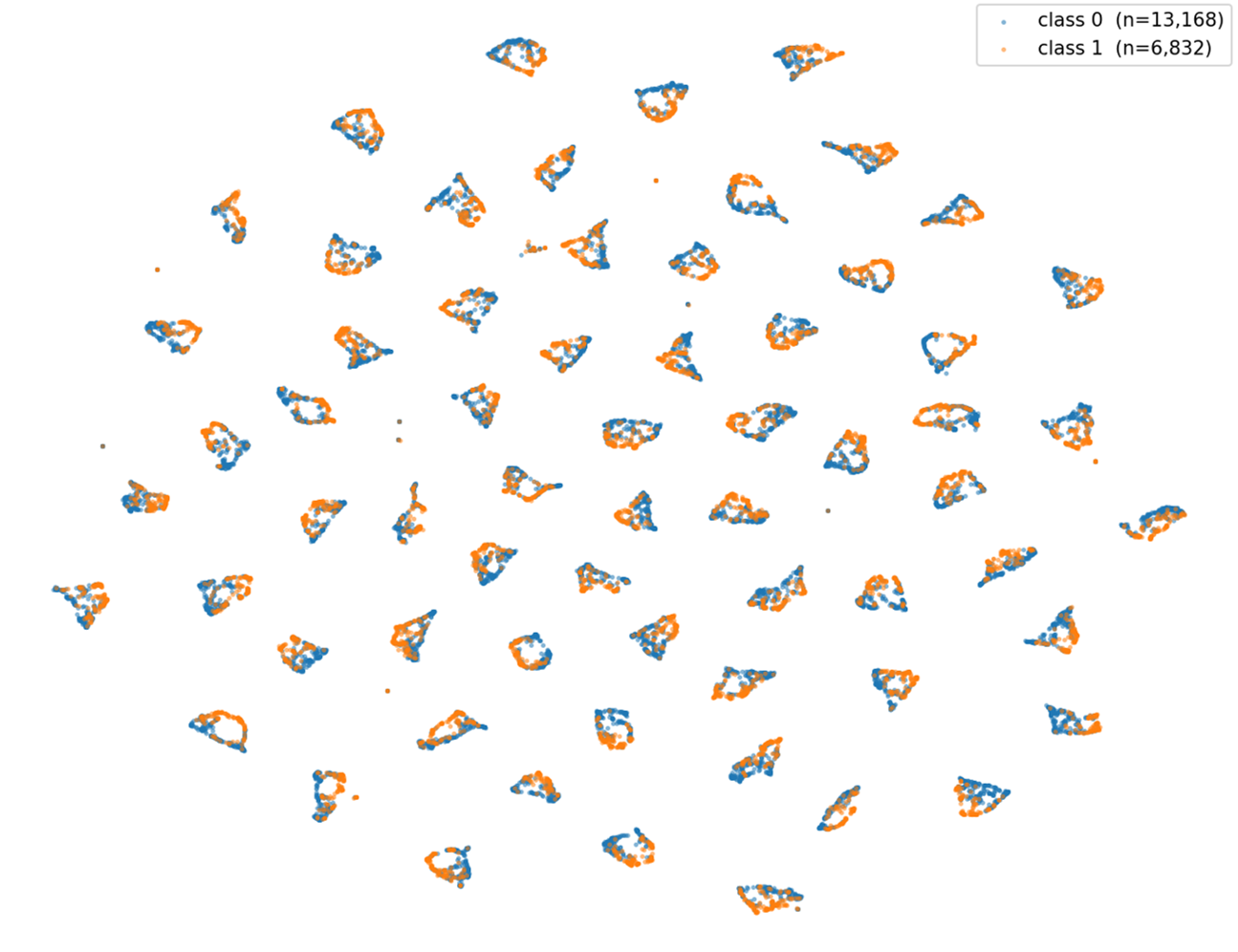}
    \caption{DINOv3-ViT-L-O-Net (natural)}
    \label{fig:umap-nat}
  \end{subfigure}
  \\[1ex]
  \begin{subfigure}[b]{0.35\textwidth}
    \includegraphics[width=\linewidth]{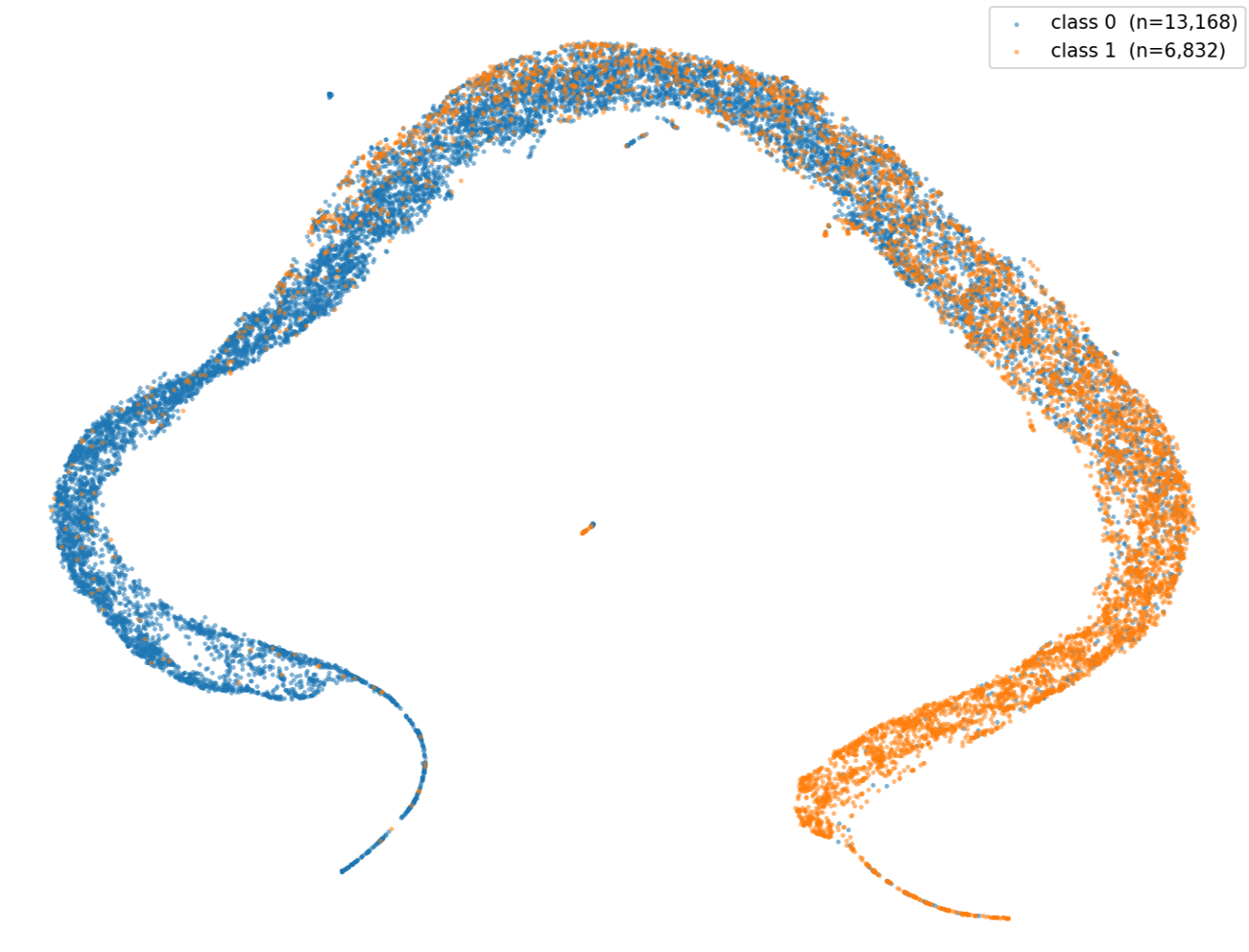}
    \caption{O-Net}
    \label{fig:umap-onet}
  \end{subfigure}
  \hspace{0.05\textwidth}
  \begin{subfigure}[b]{0.35\textwidth}
    \includegraphics[width=\linewidth]{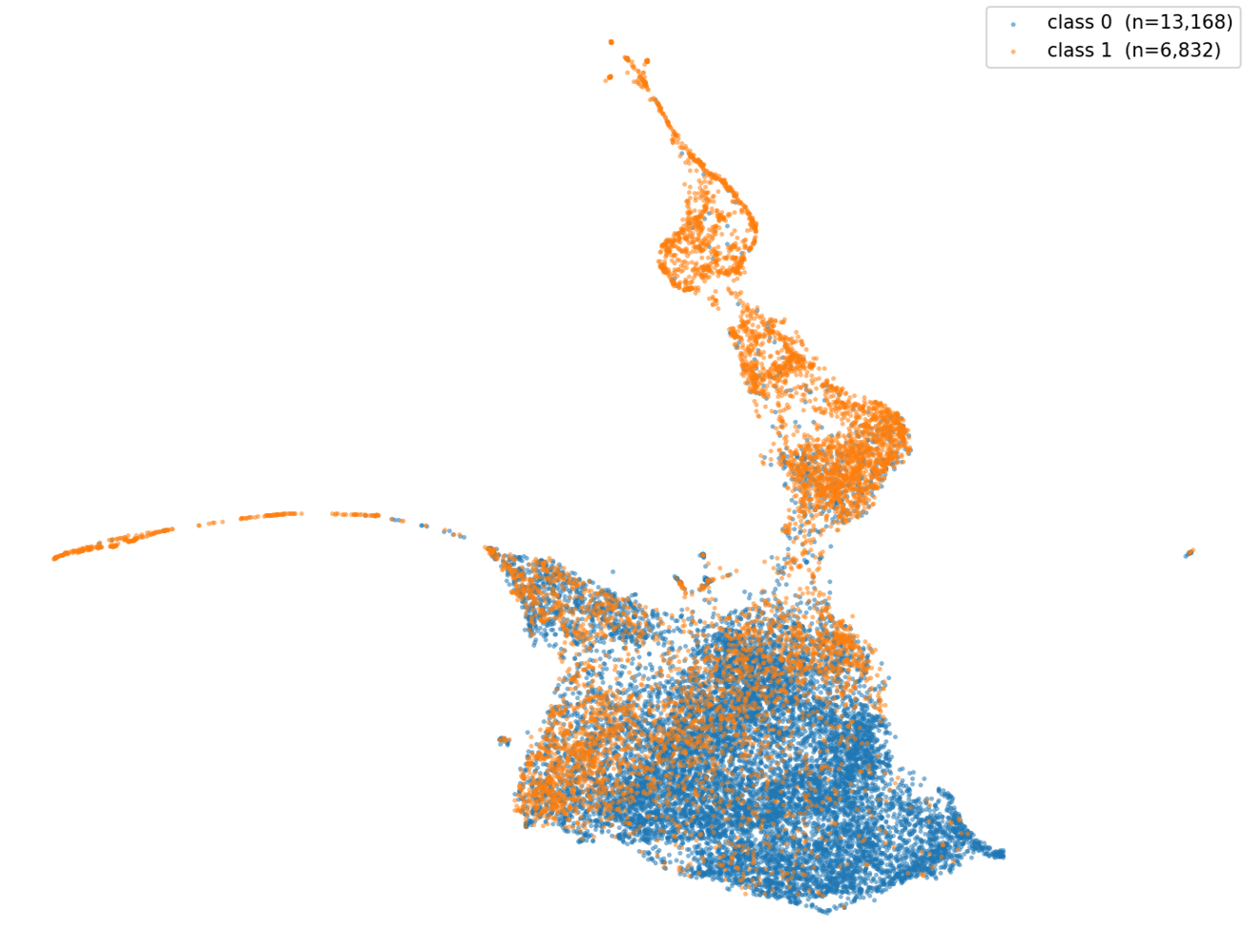}
    \caption{O-Net++}
    \label{fig:umap-onetpp}
  \end{subfigure}
  \caption{Two-dimensional UMAP projections of the frozen per-pixel features (BYOL--JSD objective) for the four backbones, coloured by class (class~0: ice/background, $n{=}13{,}168$; class~1: crevasse/foreground, $n{=}6{,}832$). The DINOv3 backbones (a,~b) fragment pixels into many small clusters in which the two classes are locally interleaved, whereas the convolutional backbones (c,~d) embed pixels onto a single connected manifold whose ends are dominated by different classes.}
  \label{fig:umap-comparison}
\end{figure*}

\begin{table}[tb]
\centering
\tablestyle
\caption{Layer-wise probing of DINOv3-ViT-L-Sat-O-Net-BYOL-JSD decoder blocks and the frozen DINOv3-ViT-L-Sat encoder, on upsampled features against pixel-level labels.}
\label{tab:layerwise}
\begin{tabular}{@{}l rr @{\hspace{12pt}} rr@{}}
\toprule
 & \multicolumn{2}{c}{Linear} & \multicolumn{2}{c}{XGBoost} \\
\cmidrule(lr){2-3}\cmidrule(l){4-5}
Features & mDSC & mIoU & mDSC & mIoU \\
\midrule
Decoder block 1       & 71.81 & 57.64 & 74.47 & 60.85 \\
Decoder block 2       & 73.35 & 59.32 & 76.95 & 64.06 \\
Decoder block 3       & 48.27 & 33.87 & 74.82 & 61.18 \\
DINOv3 encoder output & 67.26 & 52.33 & 77.18 & 64.17 \\
\bottomrule
\end{tabular}
\end{table}

\begin{table*}[tb]
\centering
\tablestyle
\caption{Comparison with alternative encoders, a fully supervised model and the frozen DINOv3-ViT-L-Sat encoder output (XGBoost readout).}
\label{tab:backbones}
\begin{tabular}{@{}lrr@{}}
\toprule
Method & mDSC & mIoU \\
\midrule
DINOv3-ViT-L-Sat-O-Net-BYOL-JSD-Ensemble (ours) & 77.04 & 64.28 \\
DINOv3-ViT-L-Sat-O-Net-BYOL-JSD (ours)          & 74.75 & 61.16 \\
RVSA-MAE-ViT-L-O-Net-BYOL-JSD                   & 48.39 & 32.82 \\
DINOv3-U-Net-ViT-L Supervised                   & \textbf{77.89} & \textbf{65.18} \\
DINOv3 Encoder Output                           & 77.18 & 64.17 \\
\bottomrule
\end{tabular}
\end{table*}

\section{Qualitative results}

\subsection{Feature-space geometry}
\label{sec:umap}

To understand \emph{why} the readout choice matters so much for the DINOv3 models, \autoref{fig:umap-comparison} shows two-dimensional UMAP projections of the final-decoder-block per-pixel embeddings for the four BYOL-JSD backbones, coloured by class. O-Net (\autoref{fig:umap-onet}) embeds pixels onto a single, connected S-shaped manifold with ice concentrated on one arm and crevasse on the other, and O-Net++ (\autoref{fig:umap-onetpp}) produces a comparable elongated manifold with ice on the left and crevasses on the right: a single global direction orders most of the two classes, the structure a linear boundary exploits and consistent with the convolutional backbones' strongest linear results (BYOL, JSD in \autoref{tab:ssl}). However, for BYOL-JSD itself, both convolutional backbones score poorly under either readout, so a class-ordered 2-D projection is not sufficient for separability in the original feature space.

The DINOv3 variants differ markedly. Both the satellite (\autoref{fig:umap-sat}) and natural-image (\autoref{fig:umap-nat}) models fragment the embeddings into dozens of small, disconnected clusters — thin filaments for the satellite checkpoint and compact blobs for the natural-image checkpoint — within which the two classes are locally interleaved rather than globally sorted. No single linear direction separates ice from crevasse across these clusters, so the linear probe collapses to a constant prediction, whereas a partition-based classifier such as XGBoost can carve each local cluster independently, which is why the same features rise to the top under the non-linear readout. As UMAP is itself a non-linear embedding, these projections illustrate rather than prove (non-)separability in the original feature space.

This geometry should not be read as evidence that the crevasse-trained O-Net is the better \emph{encoder}, and it is not a property of DINOv3 features per se: as \autoref{sec:layers} shows, the frozen DINOv3-Sat encoder output is substantially linearly separable (67.26 mDSC), and the fragmentation arises in the final decoder block. The O-Net versus DINOv3-O-Net comparison also varies training-data domain, frozen versus trainable encoder, and architecture simultaneously, so the difference cannot be attributed to any one factor.

\subsection{Decoder depth, frozen encoder and qualitative results}
\label{sec:layers}

\begin{figure}[tb]
  \centering
  \captionsetup[subfigure]{justification=centering}
  \begin{subfigure}[t]{0.31\columnwidth}
    \centering
    \includegraphics[width=\linewidth]{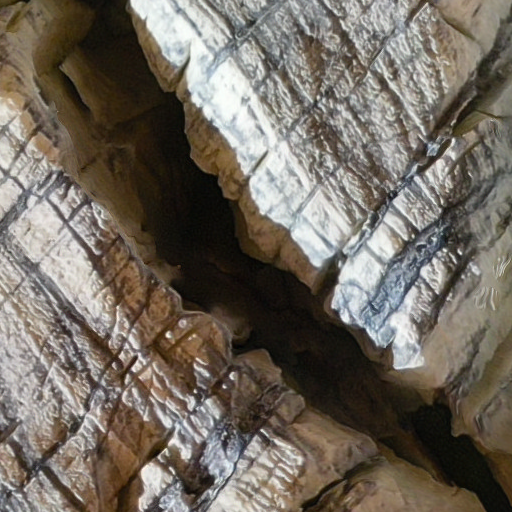}
    \caption{Input Tile}
    \label{fig:natural_a}
  \end{subfigure}%
  \hfill
  \begin{subfigure}[t]{0.31\columnwidth}
    \centering
    \includegraphics[width=\linewidth]{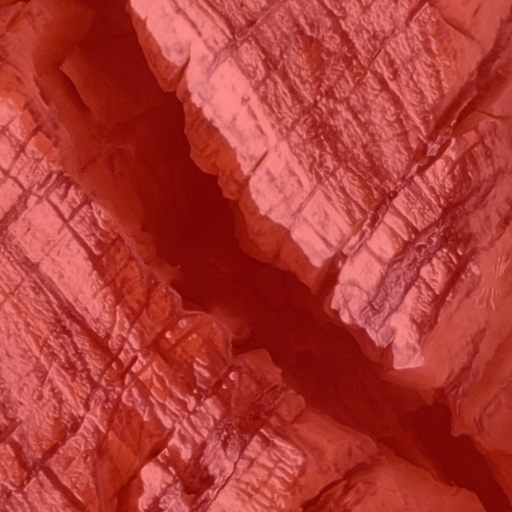}
    \caption{Linear-Probe}
    \label{fig:natural_b}
  \end{subfigure}%
  \hfill
  \begin{subfigure}[t]{0.31\columnwidth}
    \centering
    \includegraphics[width=\linewidth]{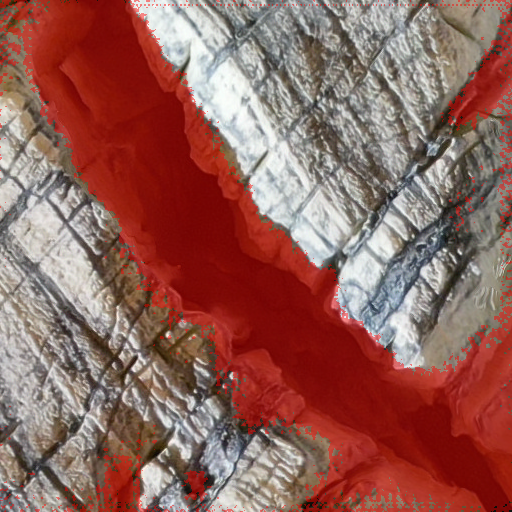}
    \caption{Ensemble Linear-Probe}
    \label{fig:natural_c}
  \end{subfigure}%
  \\[4pt]
  \begin{subfigure}[t]{0.31\columnwidth}
    \centering
    \includegraphics[width=\linewidth]{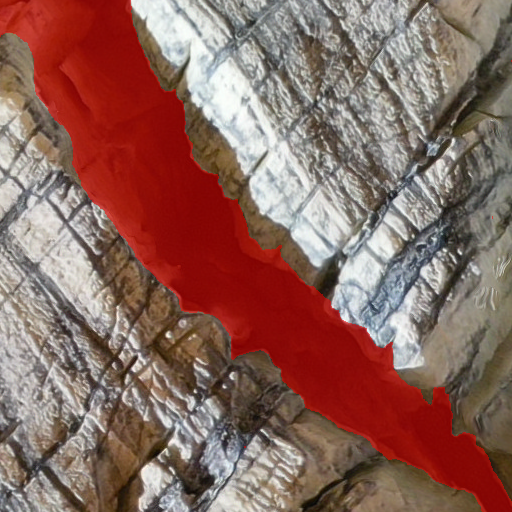}
    \caption{Ground Truth}
    \label{fig:overlay_a}
  \end{subfigure}%
  \hfill
  \begin{subfigure}[t]{0.31\columnwidth}
    \centering
    \includegraphics[width=\linewidth]{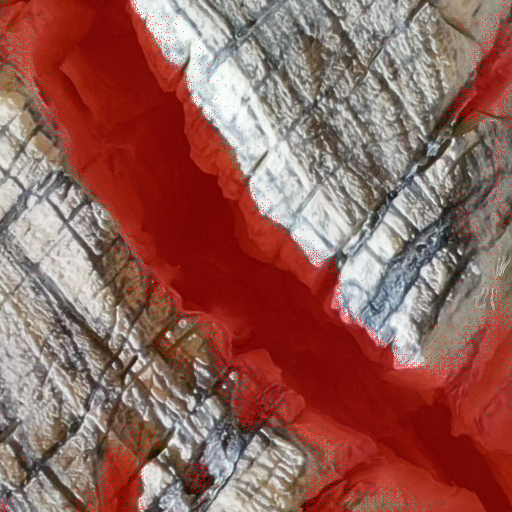}
    \caption{XGBoost}
    \label{fig:overlay_b}
  \end{subfigure}%
  \hfill
  \begin{subfigure}[t]{0.31\columnwidth}
    \centering
    \includegraphics[width=\linewidth]{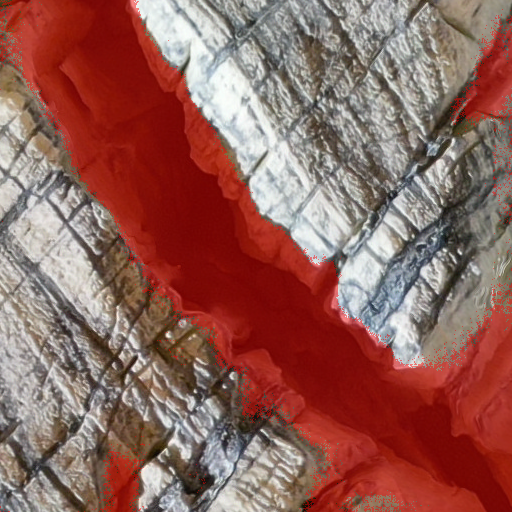}
    \caption{Ensemble XGBoost}
    \label{fig:overlay_c}
  \end{subfigure}%
  \caption{Comparison between the input tile, ground truth mask, and masks from four evaluation methods, displaying the majority-class collapse under linear evaluation for DINOv3-ViT-L-Sat-O-Net-BYOL-JSD and not the other methods.}
  \label{fig:qualitative-results-2}
\end{figure}

Layer-wise probing of DINOv3-ViT-L-Sat-O-Net-BYOL-JSD (\autoref{tab:layerwise}) localises the linear collapse to the final decoder block 3: blocks 1 and 2 remain linearly separable (71.81 and 73.35 mDSC), whereas block 3 collapses. Under XGBoost, block 2 already outperforms block 3. The ensemble's resistance to collapse is therefore attributable to blocks 1 and 2 retaining more linearly separable representations, with soft-voting diluting block 3's collapse. \autoref{fig:qualitative-results-2} shows the collapse for a single test tile, and \autoref{fig:qualitative-results} in the appendix compares predictions on three test tiles against the unsupervised baselines. Collapse under linear evaluation among the methods in (\autoref{tab:dl}), occurs only for the final-block features of our model, not for the ensemble nor for any of the five unsupervised baselines. We hypothesise that the baselines are less prone to collapse because their correspondence-based objectives use negative samples that push apart representations of different pixels, whereas the non-contrastive BYOL-JSD objective relies on stop-gradient and a predictor network alone.

Most notably, the frozen DINOv3-Sat encoder output, read out directly with XGBoost without SSL fine-tuning or a decoder, is the strongest of our label-efficient configurations (\autoref{tab:layerwise}). It reaches 77.18 mDSC / 64.17 mIoU, effectively level with our ensemble (77.04 mDSC) and above every individual decoder block, 0.71 mDSC behind the fully supervised DINOv3-ViT-L U-Net (77.89 mDSC; \autoref{tab:backbones}) and 1.36 mDSC behind HP (78.54 mDSC). Under the linear probe the picture differs: the encoder output (67.26 mDSC) trails decoder blocks 1 and 2 and the ensemble (74.09 mDSC), so the intermediate decoder blocks are more linearly separable than the encoder features, even though the final block collapses. Replacing DINOv3-Sat with a remote-sensing MAE ViT-L encoder (RVSA) \cite{he2022masked, wang2022advancing} in the same pipeline drops XGBoost performance to 48.39 mDSC, close to the constant-prediction level. On this benchmark, therefore, the SSL-trained decoder adds nothing under a non-linear readout, and the frozen satellite foundation model on its own is a strong label-efficient baseline.

\section{Limitations and future work}
\label{sec:limitations}

CrevasseSeg is single-glacier data captured across two campaigns (August 2023 and September 2025), so generalisation across seasons and years remains untested. The spatial buffer is derived from image-texture autocorrelation and removes short-range leakage between splits, but all splits still share one glacier, so the benchmark measures within-site rather than cross-site generalisation. Differences among the top configurations are small and based on single-seed point estimates; multi-seed training with confidence intervals and significance testing is needed before drawing objective-level conclusions. The task is also binary, whereas crevasses are inherently multi-scale; extending CrevasseSeg to multi-class or instance-level delineation, and to satellite or multi-modal data, would broaden its value. Finally, generalising across glacier environments remains critical for automated mapping and for adapting outputs to both large-scale surveys and real-time monitoring during glacier travel.

\section{Conclusion}

We introduced CrevasseSeg, a label-efficient benchmark for crevasse segmentation from UAV orthomosaic imagery with spatially buffered splits derived from the measured autocorrelation range of the imagery, and used it to benchmark five SSL objectives across three architectures (O-Net, O-Net++, DINOv3-O-Net) under linear and non-linear readouts. At the final decoder block, DINOv3-initialised models are mostly the weakest under linear probing yet the strongest under a non-linear readout, and satellite pre-training helps for every objective under the non-linear readout. A simple pipeline of SSL fine-tuning plus an XGBoost readout on 24 labels outperforms standard baselines, reaching 74.75 mDSC / 61.16 mIoU for a single model and 77.04 mDSC / 64.28 mIoU for a decoder-block soft-voting ensemble that also removes linear-probe collapse. 
Yet the frozen DINOv3-Sat encoder alone, read out with XGBoost, reaches 77.18 mDSC / 64.17 mIoU, within 0.71 mDSC of a fully supervised DINOv3 U-Net, showing that off-the-shelf satellite foundation models are the baseline to beat. We hope CrevasseSeg supports further work on label-efficient segmentation in remote sensing, particularly evaluation protocols that do not rely on linear separability alone.

\section*{Acknowledgments}

This research was supported by funding secured from the University's Development Trust. We acknowledge the generous support of alumni and friends in establishing the University of Aberdeen's Interdisciplinary Institute, which enabled this research, including Dr Jane Hellman Caseley (MBChB 1956), Professor Patrick Meares (DSc 1959), Nancy Miller (MA 1942), Norman Robertson, Dr Ian Slessor (MBChB 1956) and Anne Young (MA 1957).

Fieldwork was led by W.D. Harcourt, and UAV drone data acquisition in Svalbard was led by R. Hann. The authors would like to thank Danni Pearce and Wojciech Gajek for further assistance in acquiring the drone data in Svalbard that formed the basis of the data set used in this study.

\section*{LLM Disclosure}

In this work, we used Claude Opus models (Anthropic; the versions available on claude.ai between August and November 2026), accessed through the claude.ai web interface, for methodology and implementation: they assisted in designing the spatial-autocorrelation analysis used to set the buffer distance between dataset splits and the spatially buffered tile-sampling strategy (\autoref{sec:buffer}), and in writing the Python scripts that compute the autocorrelation buffer and generate the buffered train, validation and test tiles. We have not used LLMs/AI tools for theoretical or conceptual frameworks, mathematical claims or proofs, hypotheses, data cleaning or interpreting results; synthetic data, translation and qualitative data analysis are not applicable to this work. Additionally, we used Claude to assist in creating figures (the autocorrelation profile plot in \autoref{fig:autocorr}, produced by the analysis script, and guidance on producing the region-of-interest map in \autoref{fig:aoi-both}), to edit the paper to improve readability, and to check that the results reported in the text matched those in the tables. We have reviewed all AI-assisted work: the authors ran and checked the scripts and their outputs, reviewed the spatial-autocorrelation method and buffer distance, checked all suggested text edits against the original draft, and verified all reported results against the original experimental outputs. We take responsibility for the final content of this work, including text, claims, or artefacts produced with the aid of LLMs/AI tools.

\printbibliography

\appendix

\clearpage
\onecolumn
\section{Borebreen QGIS orthomosaic}

\begin{figure}[H]  
  \centering
  \setlength{\tabcolsep}{1pt}
  \renewcommand{\arraystretch}{0}
  \begin{tabular}{c}
    \footnotesize Borebreen -- 8th August 2023 \\ [3pt]
    \includegraphics[width=0.65\textwidth,height=0.36\textheight,keepaspectratio]{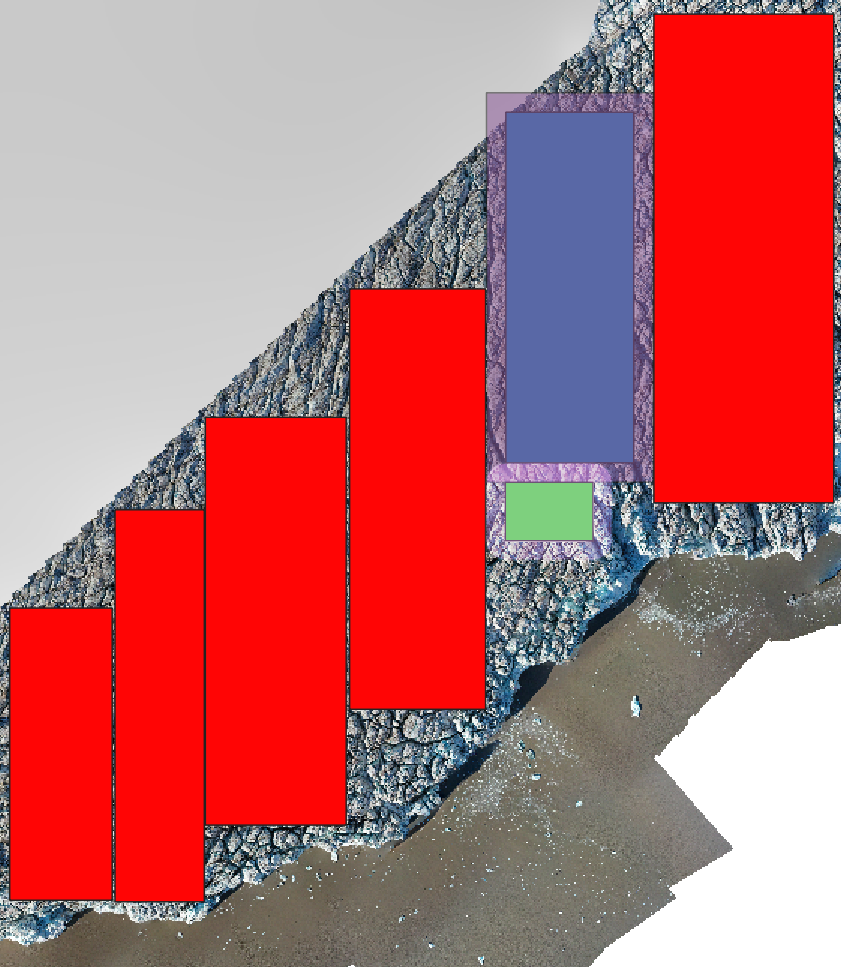} \\ [6pt]

    \footnotesize Borebreen -- 8th September 2025 \\ [3pt]
    \includegraphics[width=0.65\textwidth,height=0.36\textheight,keepaspectratio]{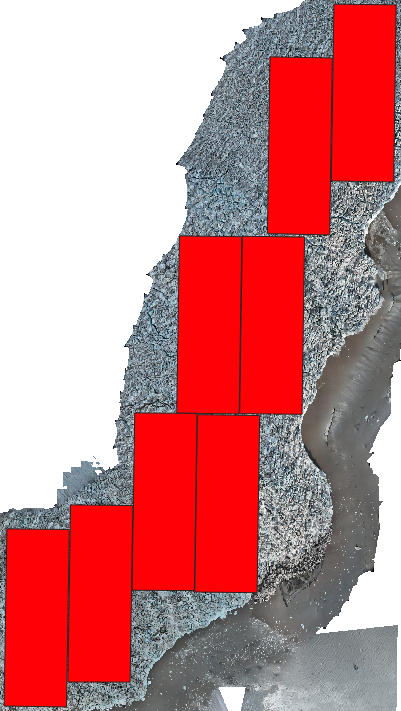} \\ [6pt]
  \end{tabular}

  \vspace{4pt}
  \definecolor{trainred}{RGB}{237, 28, 36}
  \definecolor{valgreen}{RGB}{34, 139, 34}
  \definecolor{testblue}{RGB}{31, 119, 180}
  \colorbox{trainred}{\rule{0pt}{6pt}\rule{6pt}{0pt}} \, \footnotesize Train \qquad
  \colorbox{valgreen}{\rule{0pt}{6pt}\rule{6pt}{0pt}} \, \footnotesize Validation \qquad
  \colorbox{testblue}{\rule{0pt}{6pt}\rule{6pt}{0pt}} \, \footnotesize Test

  \caption{Regions of interest used for training, validation, and testing at Borebreen on 8th August 2023 (top) and 8th September 2025 (bottom).}
  \label{fig:aoi-both}
\end{figure}







\begin{figure}[p]
  \centering
  \setlength{\qimg}{0.128\textwidth}
  \setlength{\tabcolsep}{0.75pt}
  \renewcommand{\arraystretch}{0}
  \begin{tabular}{@{}c ccc @{\hspace{6pt}} c ccc@{}}
    & \multicolumn{3}{c}{\small Linear-Probe} && \multicolumn{3}{c}{\small XGBoost} \\[3pt]
    \cmidrule(r){2-4}\cmidrule(l){6-8} \\[-1pt]
    \qlabel{Image} &
      \includegraphics[width=\qimg]{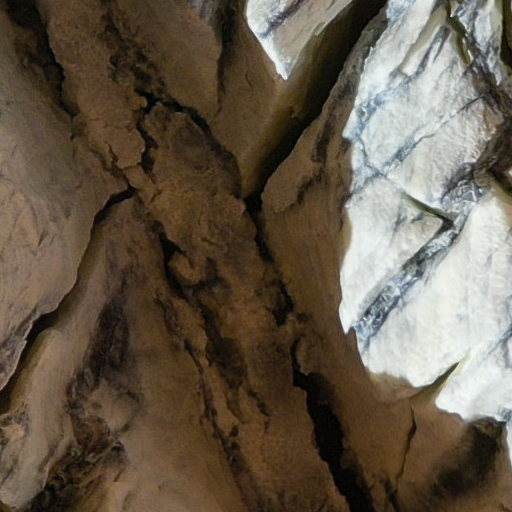} & \includegraphics[width=\qimg]{results/image_borebreen_50_1_1.png} & \includegraphics[width=\qimg]{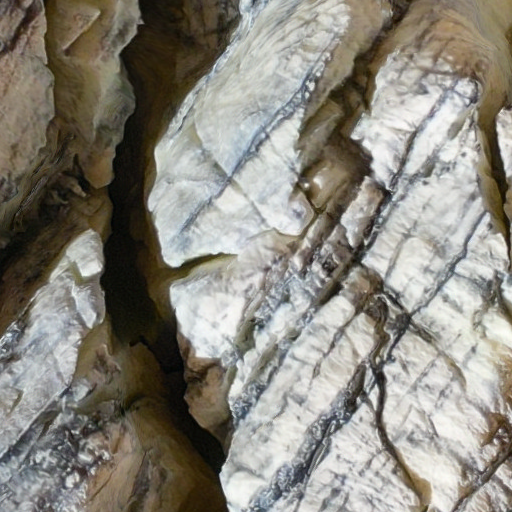} &&
      \includegraphics[width=\qimg]{results/image_borebreen_7_0_0.png} & \includegraphics[width=\qimg]{results/image_borebreen_50_1_1.png} & \includegraphics[width=\qimg]{results/image_borebreen_27_0_1.png} \\[1.5pt]
    \qlabel{GT} &
      \includegraphics[width=\qimg]{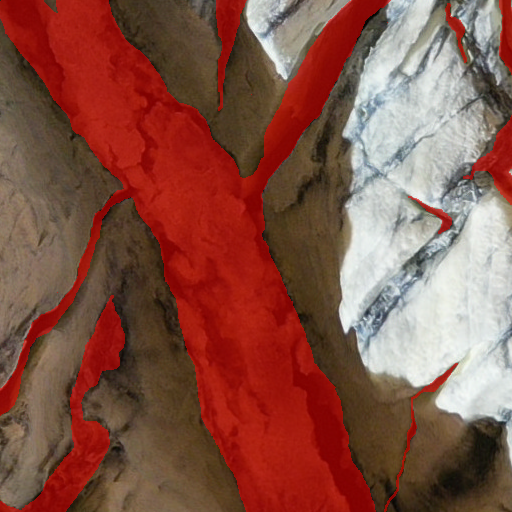} & \includegraphics[width=\qimg]{results/image_borebreen_50_1_1_overlay.png} & \includegraphics[width=\qimg]{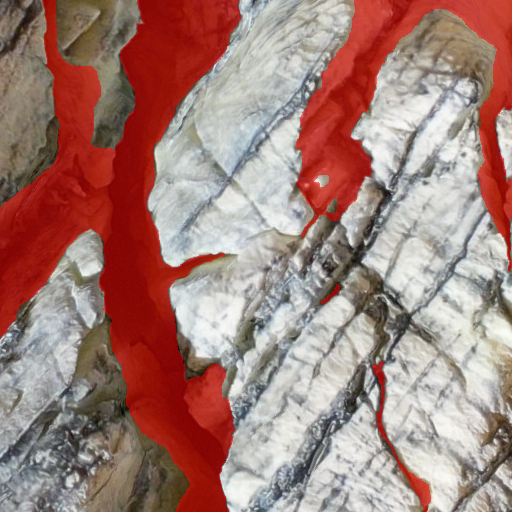} &&
      \includegraphics[width=\qimg]{results/image_borebreen_7_0_0_overlay.png} & \includegraphics[width=\qimg]{results/image_borebreen_50_1_1_overlay.png} & \includegraphics[width=\qimg]{results/image_borebreen_27_0_1_overlay.png} \\[1.5pt]
    \qlabel{PriMaPs} &
      \includegraphics[width=\qimg]{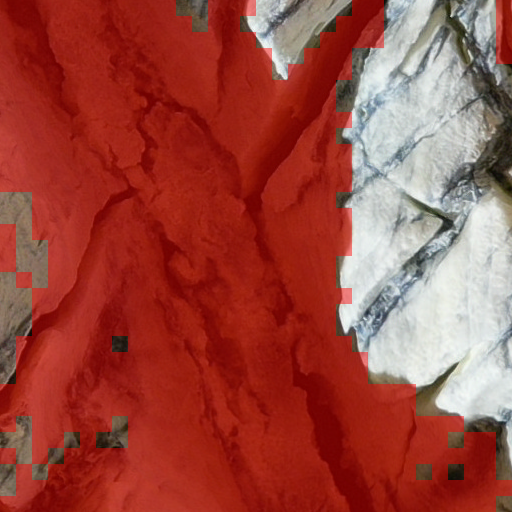} & \includegraphics[width=\qimg]{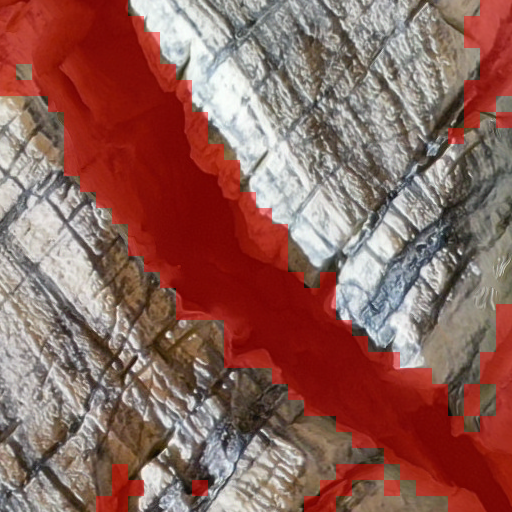} & \includegraphics[width=\qimg]{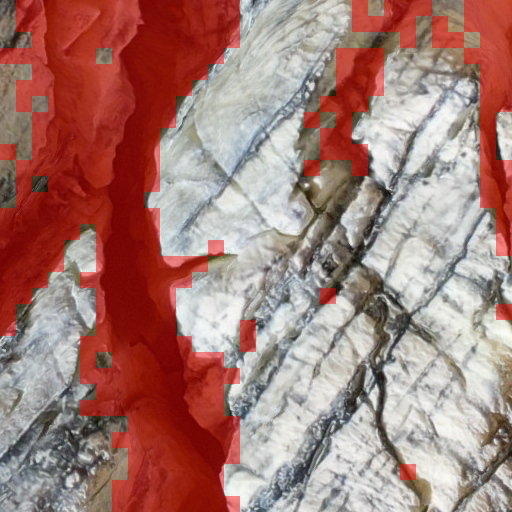} &&
      \includegraphics[width=\qimg]{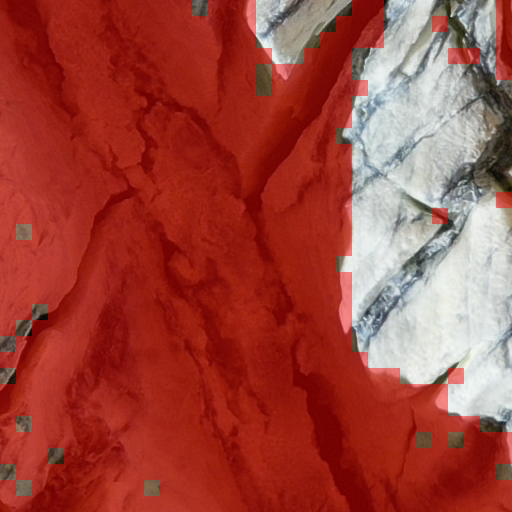} & \includegraphics[width=\qimg]{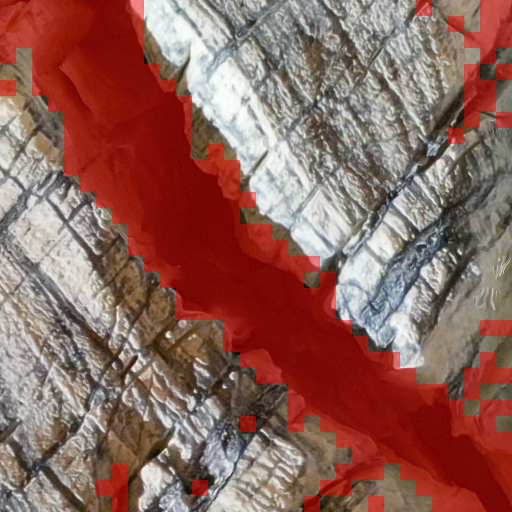} & \includegraphics[width=\qimg]{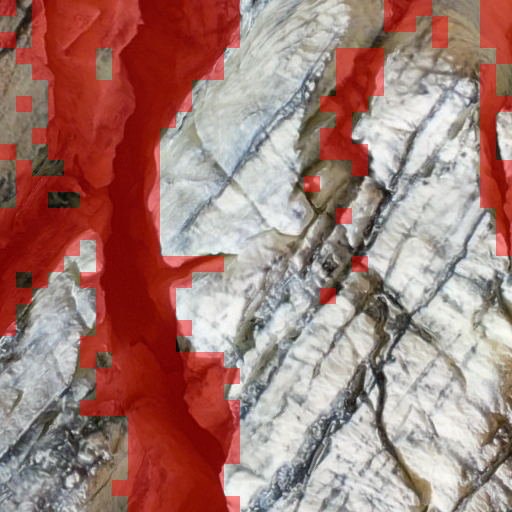} \\[1.5pt]
    \qlabel{EAGLE} &
      \includegraphics[width=\qimg]{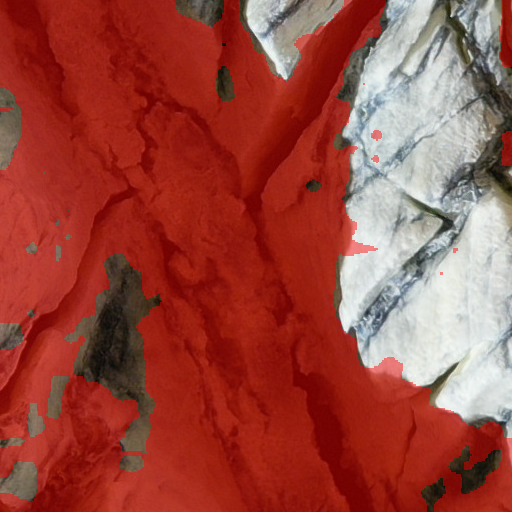} & \includegraphics[width=\qimg]{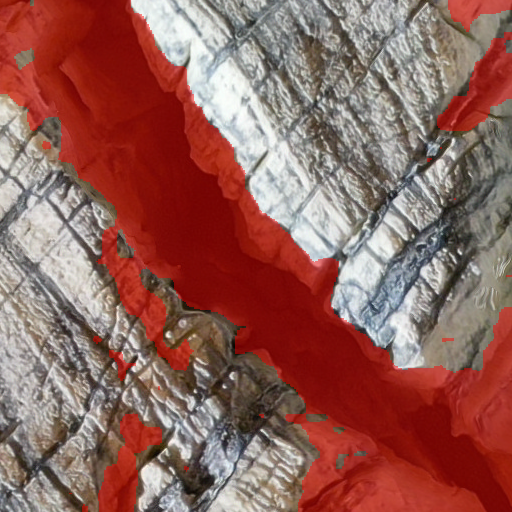} & \includegraphics[width=\qimg]{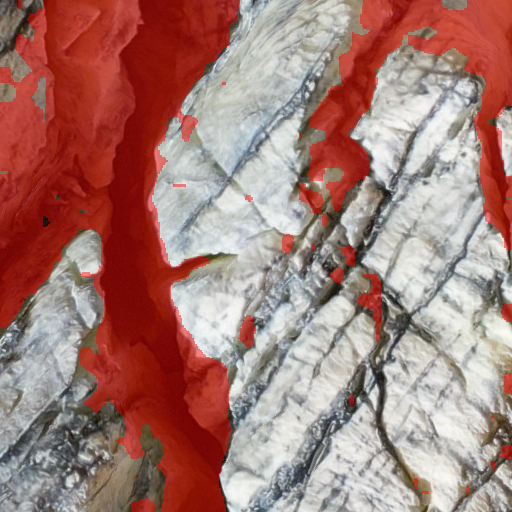} &&
      \includegraphics[width=\qimg]{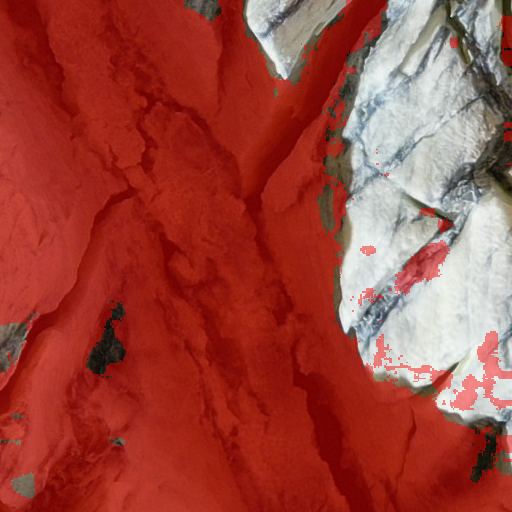} & \includegraphics[width=\qimg]{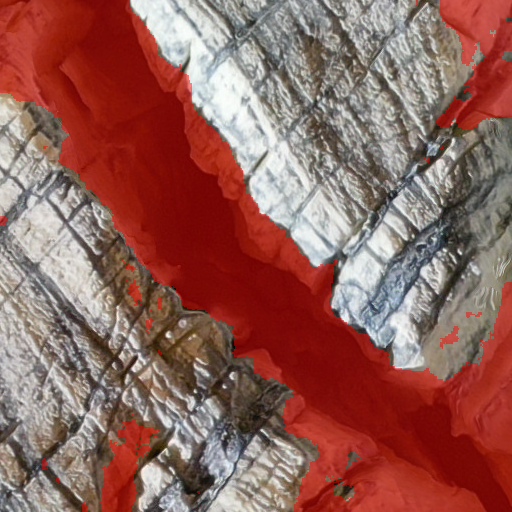} & \includegraphics[width=\qimg]{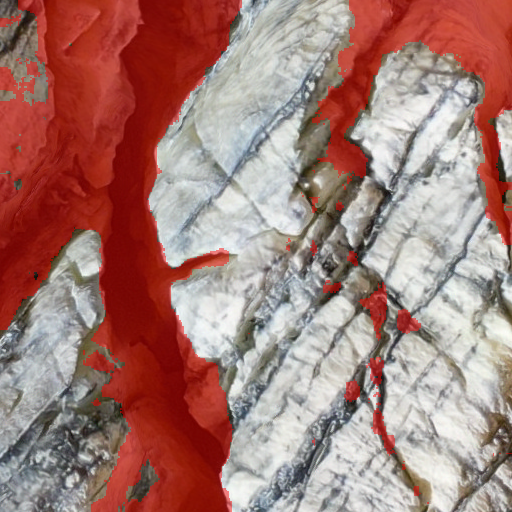} \\[1.5pt]
    \qlabel{HP} &
      \includegraphics[width=\qimg]{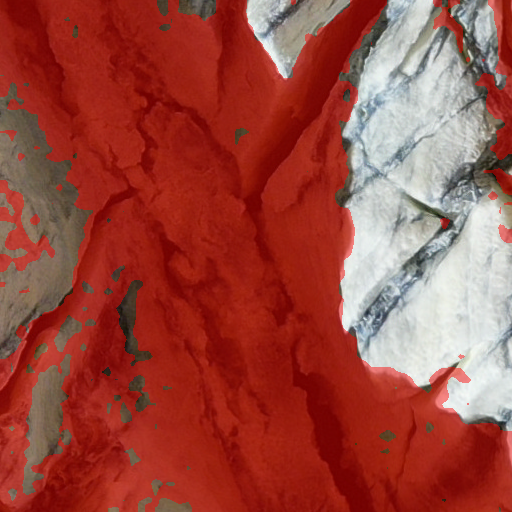} & \includegraphics[width=\qimg]{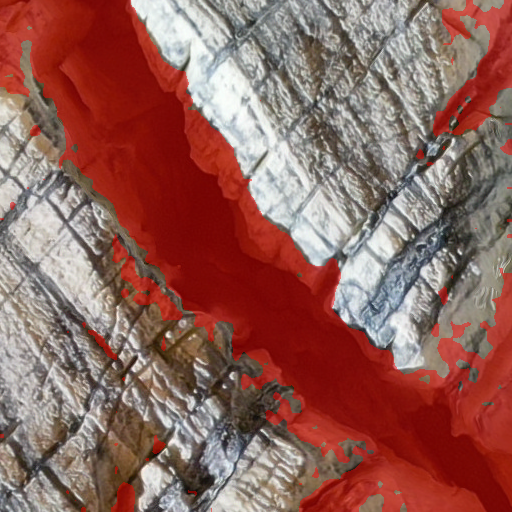} & \includegraphics[width=\qimg]{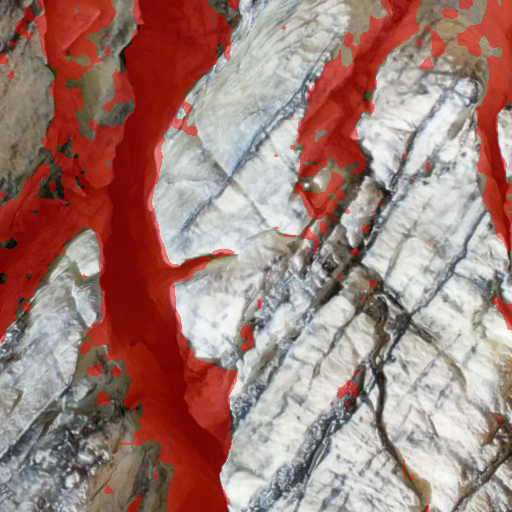} &&
      \includegraphics[width=\qimg]{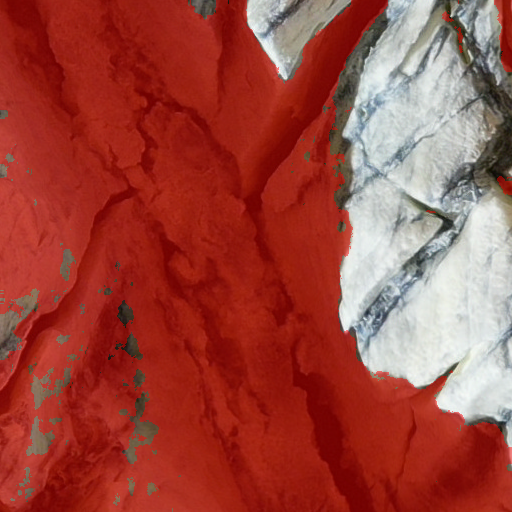} & \includegraphics[width=\qimg]{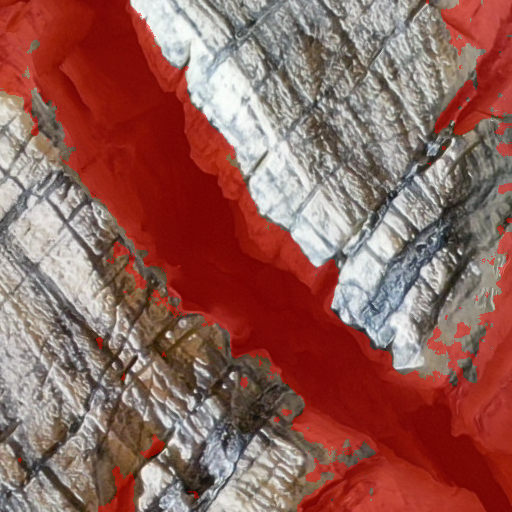} & \includegraphics[width=\qimg]{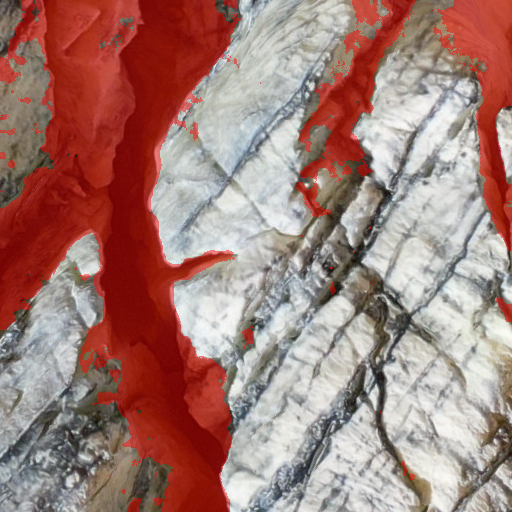} \\[1.5pt]
    \qlabel{STEGO} &
      \includegraphics[width=\qimg]{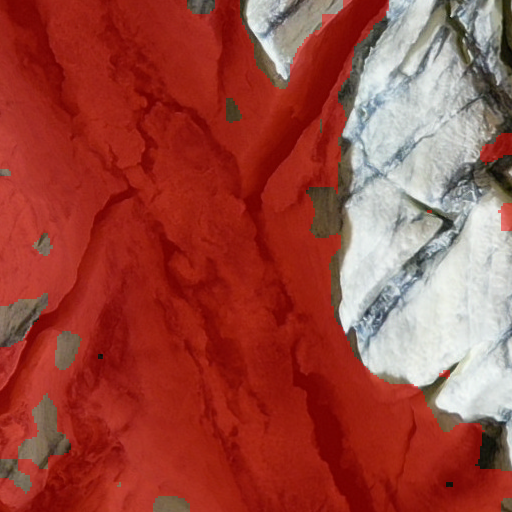} & \includegraphics[width=\qimg]{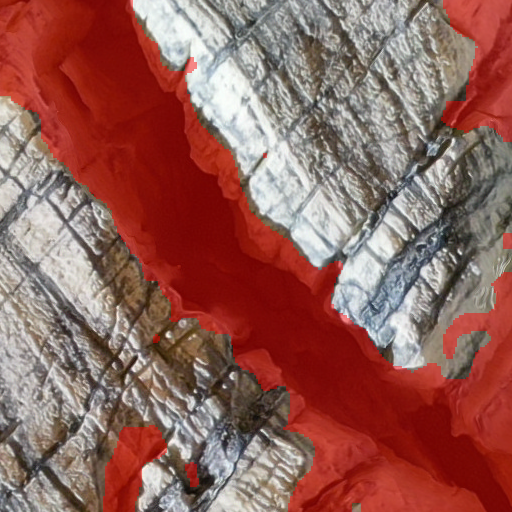} & \includegraphics[width=\qimg]{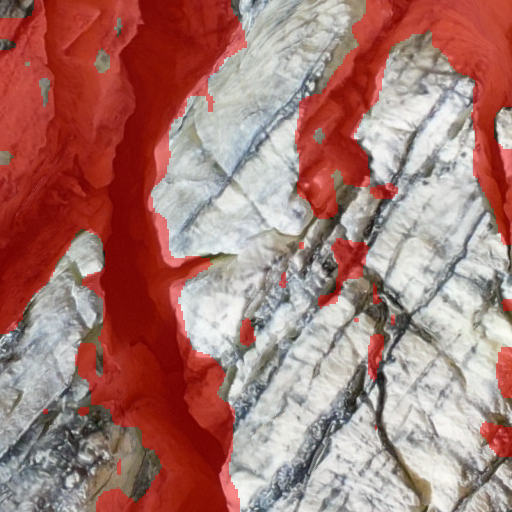} &&
      \includegraphics[width=\qimg]{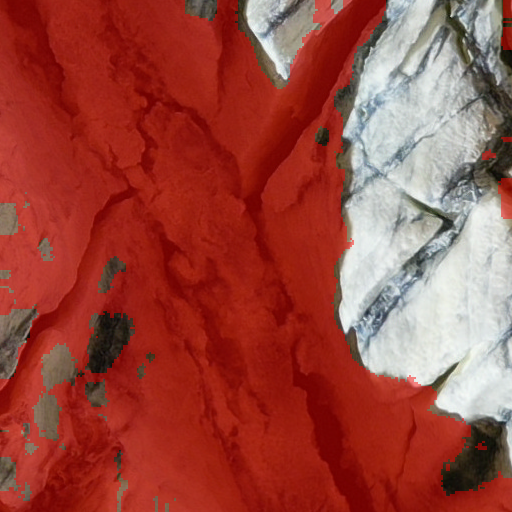} & \includegraphics[width=\qimg]{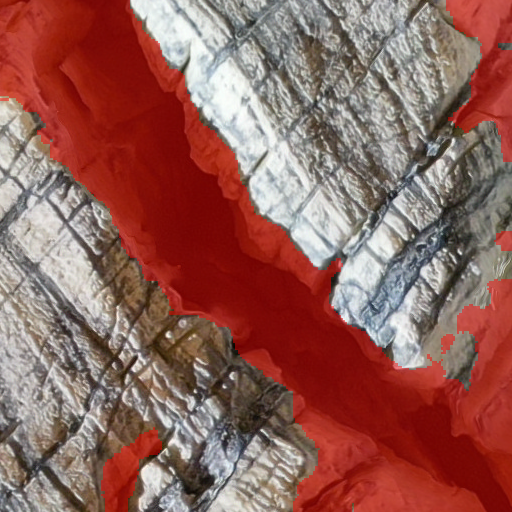} & \includegraphics[width=\qimg]{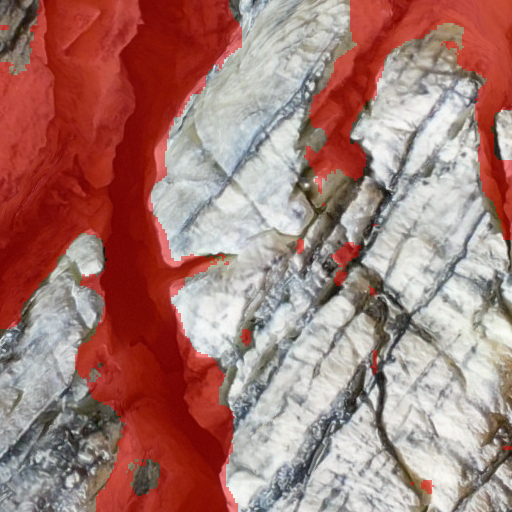} \\[1.5pt]
    \qlabel{SmooSeg} &
      \includegraphics[width=\qimg]{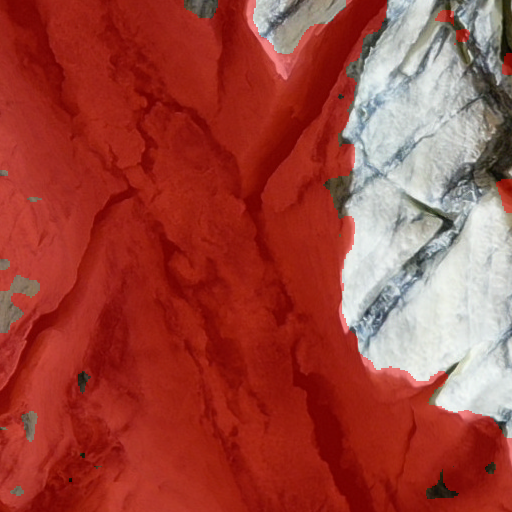} & \includegraphics[width=\qimg]{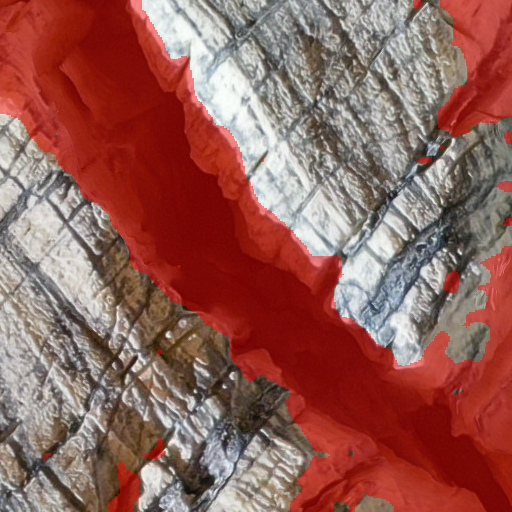} & \includegraphics[width=\qimg]{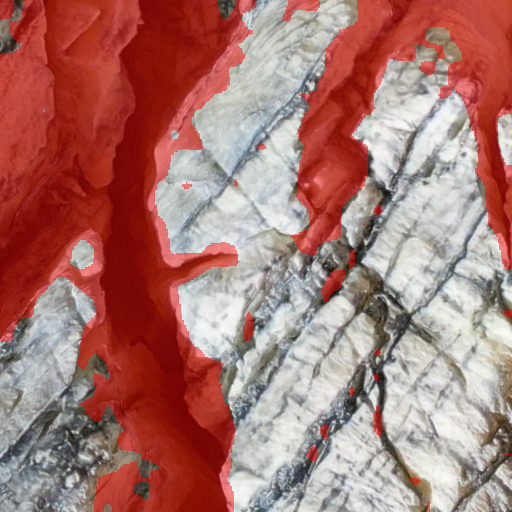} &&
      \includegraphics[width=\qimg]{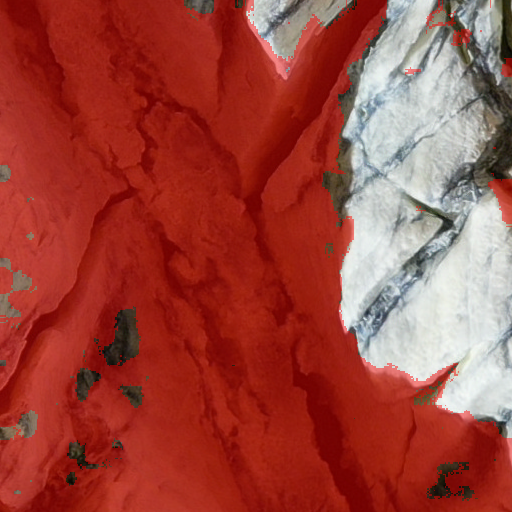} & \includegraphics[width=\qimg]{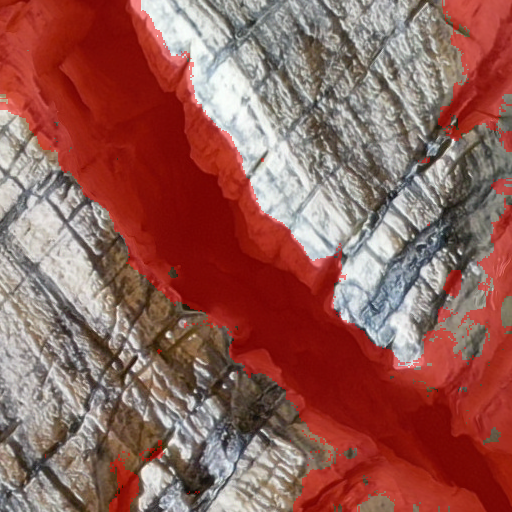} & \includegraphics[width=\qimg]{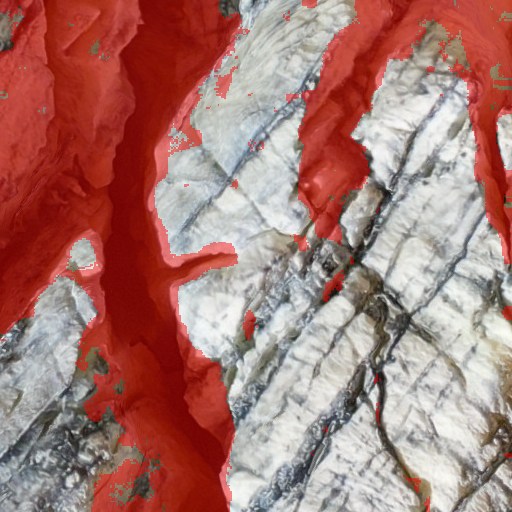} \\[1.5pt]
    \qlabel{MAE-ViT-L} &
      \includegraphics[width=\qimg]{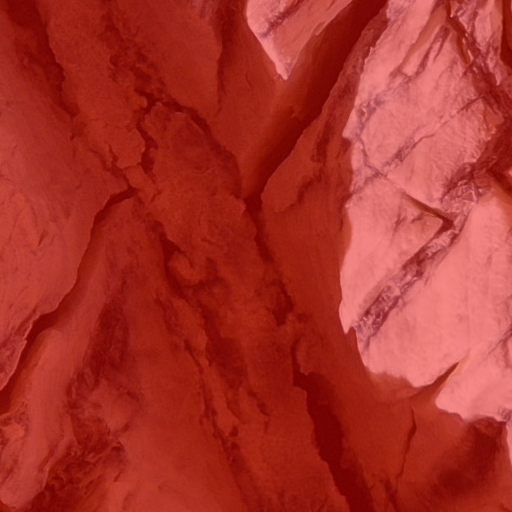} & \includegraphics[width=\qimg]{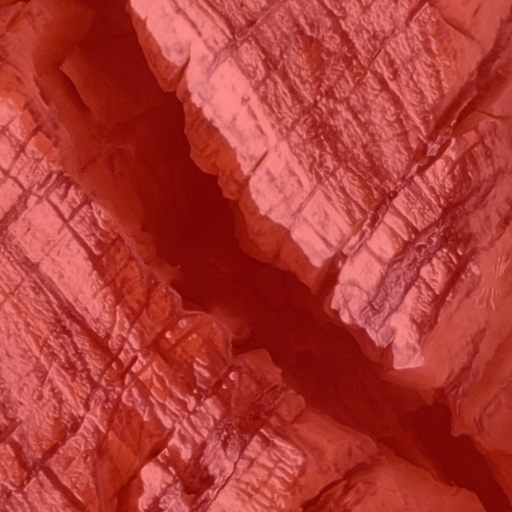} & \includegraphics[width=\qimg]{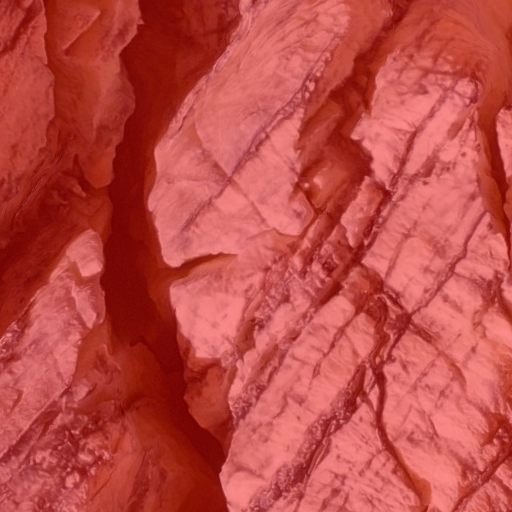} &&
      \includegraphics[width=\qimg]{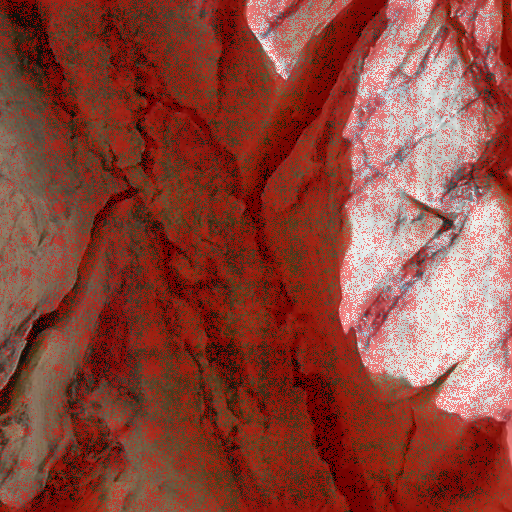} & \includegraphics[width=\qimg]{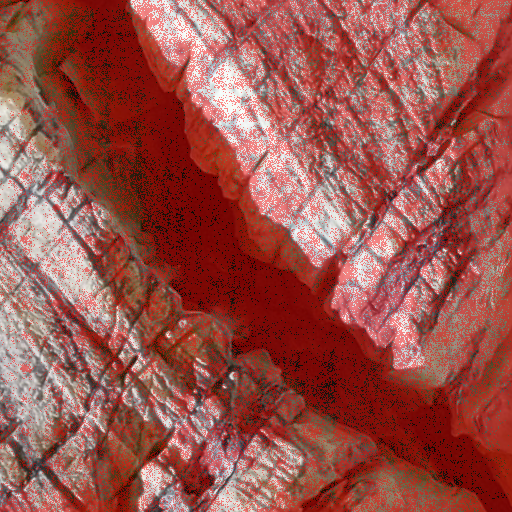} & \includegraphics[width=\qimg]{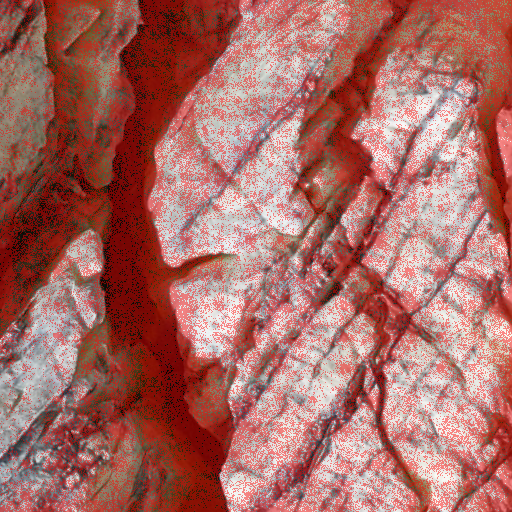} \\[1.5pt]
    \qlabel{DINOv3-Sat-BYOL-JSD-Ensemble (ours)} &
      \includegraphics[width=\qimg]{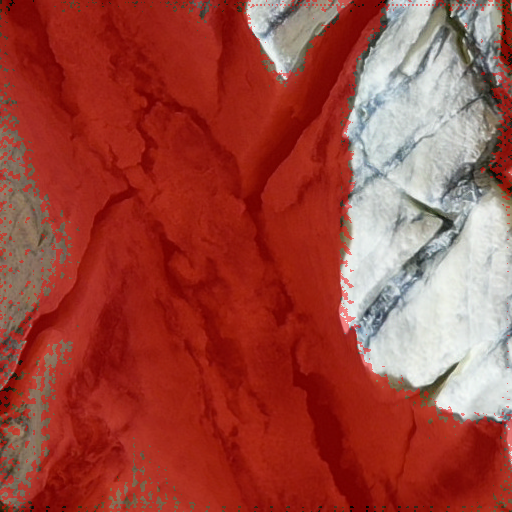} & \includegraphics[width=\qimg]{results_rebuttal_sat/Ensemble/image_borebreen_50_1_1.png} & \includegraphics[width=\qimg]{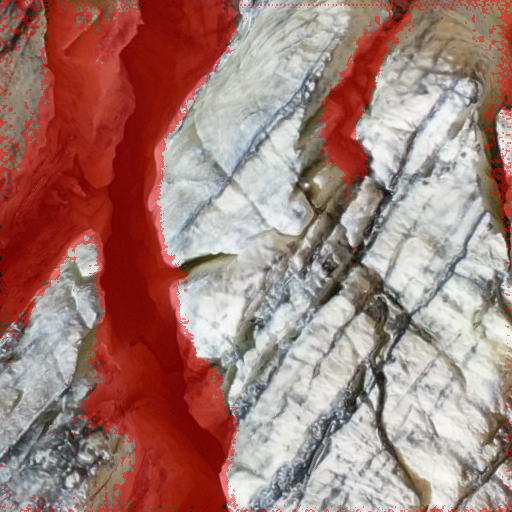} &&
      \includegraphics[width=\qimg]{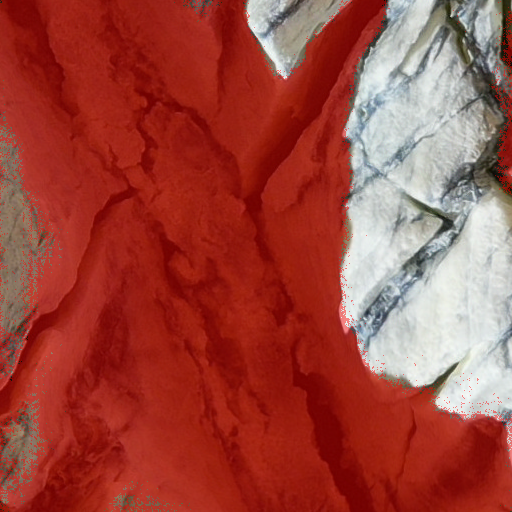} & \includegraphics[width=\qimg]{results_rebuttal_sat/Ensemble/non_image_borebreen_50_1_1.png} & \includegraphics[width=\qimg]{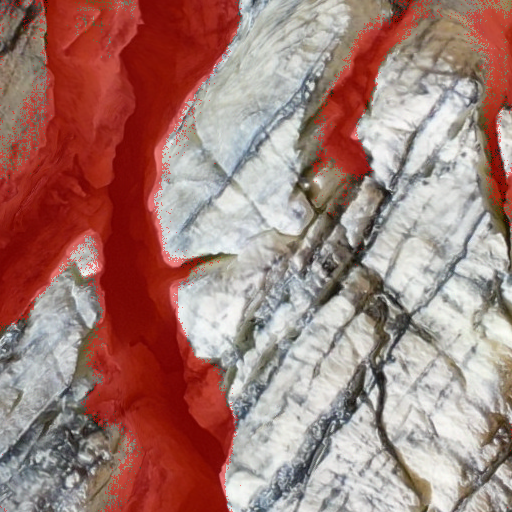} \\[1.5pt]
    \qlabel{DINOv3-Sat-BYOL-JSD (ours)} &
      \includegraphics[width=\qimg]{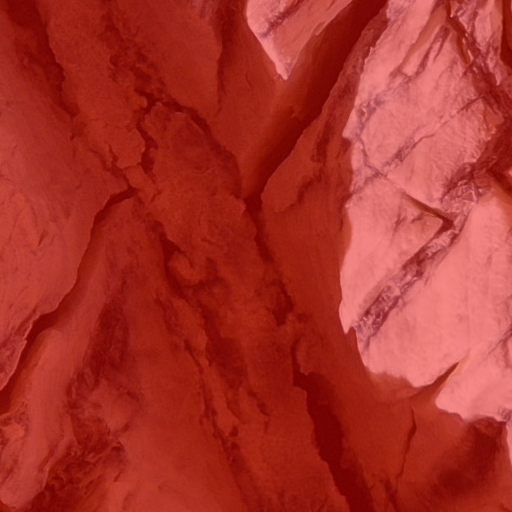} & \includegraphics[width=\qimg]{results_rebuttal_sat/lin_image_borebreen_50_1_1.png} & \includegraphics[width=\qimg]{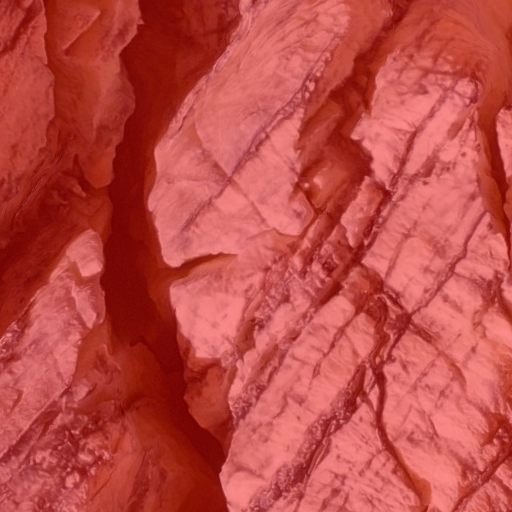} &&
      \includegraphics[width=\qimg]{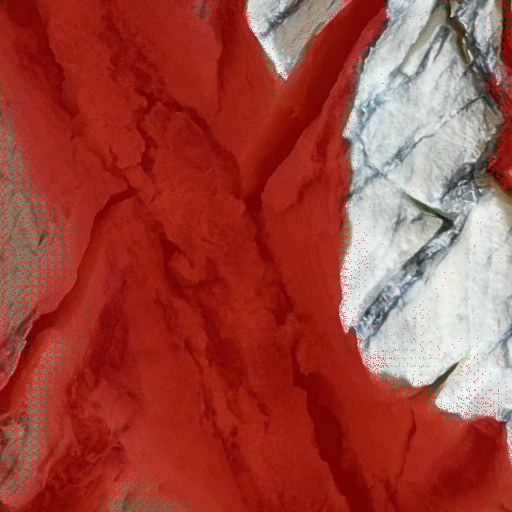} & \includegraphics[width=\qimg]{results_rebuttal_sat/non_image_borebreen_50_1_1.png} & \includegraphics[width=\qimg]{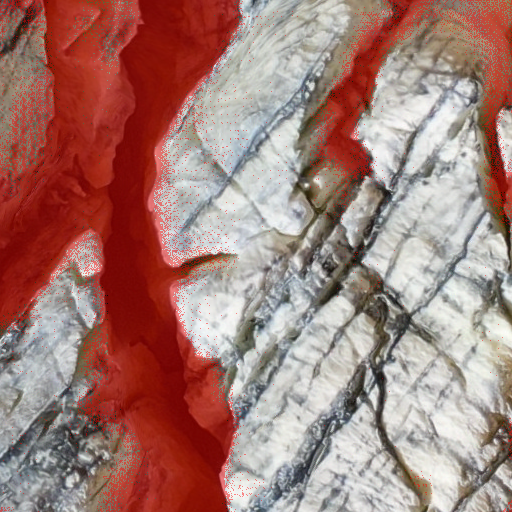} \\[1.5pt]
  \end{tabular}
  \caption{Qualitative segmentation results on three representative CrevasseSeg test tiles ($512\times512$). From top to bottom: input orthomosaic tile, ground-truth (GT) mask, and predictions from PriMaPs-EM, EAGLE, HP, STEGO, SmooSeg, DINOv3-ViT-L-Sat-O-Net-BYOL-JSD-Ensemble, and our DINOv3-ViT-L-Sat-O-Net-BYOL-JSD. Columns 1–3 use the Linear-Probe head, columns 4–6 use the XGBoost head.}
  \label{fig:qualitative-results}
\end{figure}
\twocolumn


\end{document}